\documentclass{article}
\PassOptionsToPackage{numbers,sort&compress}{natbib}

\usepackage[preprint]{neurips_2026}

\usepackage[utf8]{inputenc} % allow utf-8 input
\usepackage[T1]{fontenc}    % use 8-bit T1 fonts
\usepackage{hyperref}       % hyperlinks
\usepackage{url}            % simple URL typesetting
\usepackage{booktabs}       % professional-quality tables
\usepackage{amsfonts}       % blackboard math symbols
\usepackage{nicefrac}       % compact symbols for 1/2, etc.
\usepackage{microtype}      % microtypography
\usepackage{xcolor}         % colors

\usepackage{graphicx}
\usepackage{booktabs}
\usepackage{makecell}
\usepackage{enumitem}
\usepackage{array}
\usepackage{tabularx}
\usepackage{amsmath}
\usepackage{multirow}
\usepackage{tikz}
\usepackage{graphicx}
\usetikzlibrary{positioning, arrows.meta, shapes, fit}

\usetikzlibrary{arrows.meta,calc,positioning,decorations.pathmorphing}

\renewcommand{\arraystretch}{1.2} % global row spacing
\newcommand{\rowstrut}{\rule{0pt}{2.6ex}}

\title{No Epoch Like the Present: Robust Climate Emulation Requires Out-of-Distribution Generalisation}

\author{%
Bradley Stanley-Clamp \\
  Applied AI Lab\\
  University of Oxford, UK\\
  ~~~~~~~~\texttt{bradleysc@robots.ox.ac.uk}~~~~~~~~ \\
  \And
  Anson Lei \\
  Applied AI Lab \\
  University of Oxford, UK \\
  \texttt{anson@robots.ox.ac.uk} \\
  \AND
  Hannah M. Christensen \\
  Atmospheric, Oceanic and Planetary Physics \\
  University of Oxford, UK \\
  \texttt{hannah.christensen@physics.ox.ac.uk} \\
  \And
  Ingmar Posner \\
  Applied AI Lab \\
  University of Oxford, UK \\
  \texttt{ingmar@robots.ox.ac.uk} \\
}

\begin{document}

\maketitle

%-- Abstract --% 
\begin{abstract}
Climate emulation is an out-of-distribution (OOD) projection task. This is precisely the challenge where modern Machine Learning (ML) methods are most prone to failure. Consequently, while current ML emulators trained on present climate achieve high in-distribution performance, their future reliability under the inevitable distribution shifts of a changing climate remains a critical, poorly understood blind spot. Addressing this challenge requires a fundamental shift in how we understand, evaluate, and design climate emulators. In this work, we first confirm that climate change drives a statistically significant and progressively growing shift in atmospheric state distributions, rendering standard evaluation protocols insufficient. We empirically establish that seasonal variation serves as an effective proxy for these long-term climate shifts, providing access to \textit{real-world} distribution shifts without recourse to heuristics like synthetic perturbations. Motivated by this link, we introduce a novel evaluation framework that leverages seasonal shifts as a rigorous, zero-overhead testbed for emulator robustness. Our systematic characterisation confirms that current state-of-the-art hybrid-ML emulators degrade significantly under these realistic shifts. Finally, we chart a path forward by identifying compositional generalisation, the ability to form novel combinations from observed elementary components, as a principled route towards robust climate emulation. We demonstrate that physically motivated decompositions substantially improve OOD performance with only modest trade-offs against in-distribution performance, providing an avenue towards ML-driven climate emulators robust to an unknown future.
\end{abstract}

%-- Introduction --%
\section{Introduction}
\label{section/introduction}

Historically, numerical Earth system models (ESMs) have been the primary tools for understanding and predicting the climate system, underpinning critical decisions in climate policy, infrastructure planning, and risk management \citep{tebaldi_climate_2021,haasnoot_dynamic_2013,ranger_addressing_2013}. Climate projections are obtained by running these models autoregressively over multi-decadal timescales under evolving boundary conditions, such as rising greenhouse gas concentrations or changing sea surface temperatures. The emergent statistics of these long simulations underpin a wide range of downstream applications, from quantifying specific climate risks to diagnosing physical mechanisms. 

In an effort to overcome biases and uncertainties in ESMs, there is a continued endeavour towards improving approximations and explicitly resolving new physical processes \cite{stevens_dyamond_2019, bechtold_representing_2014}. However, these advancements are bottlenecked by their computational cost. With simulation ensembles necessary for informative prediction, alongside the need for specialised compute and long run times, the field is constrained not by a lack of questions or hypotheses, but by the sheer cost of experimentation \citep{schneider_earth_2017, alizadeh_advances_2022}.

% ML provides a potential solution
Machine learning (ML) emulation offers a potentially transformative solution to this bottleneck, significantly reducing computational costs whilst maintaining accurate prediction of complex, non-linear and highly dimensional interactions \cite{mooers_assessing_2021}. By efficiently optimising over large corpora of observational and simulation data, these emulators can potentially surpass their numerical counterparts in both speed and prediction fidelity \cite{alizadeh_advances_2022}.

% Additional challenges for ML emulators for climate
However, beyond standard machine learning challenges (e.g., inference performance and hyperparameter optimisation), ML emulators for climate modelling must satisfy two critical requirements: stability and robustness. To obtain climate projections, models are autoregressively rolled out over long time horizons, either coupled to a host model or in stand-alone configurations. Stability is therefore essential: small errors can compound over time, leading to divergence or simulation failure. More fundamentally, our changing climate system is inherently non-stationary, requiring climate emulators to operate in new, unknown data regimes. Without robustness to such changes, reliable climate projections are not possible.

\begin{figure}[t!]
    \centering
    \includegraphics[width=1\linewidth]{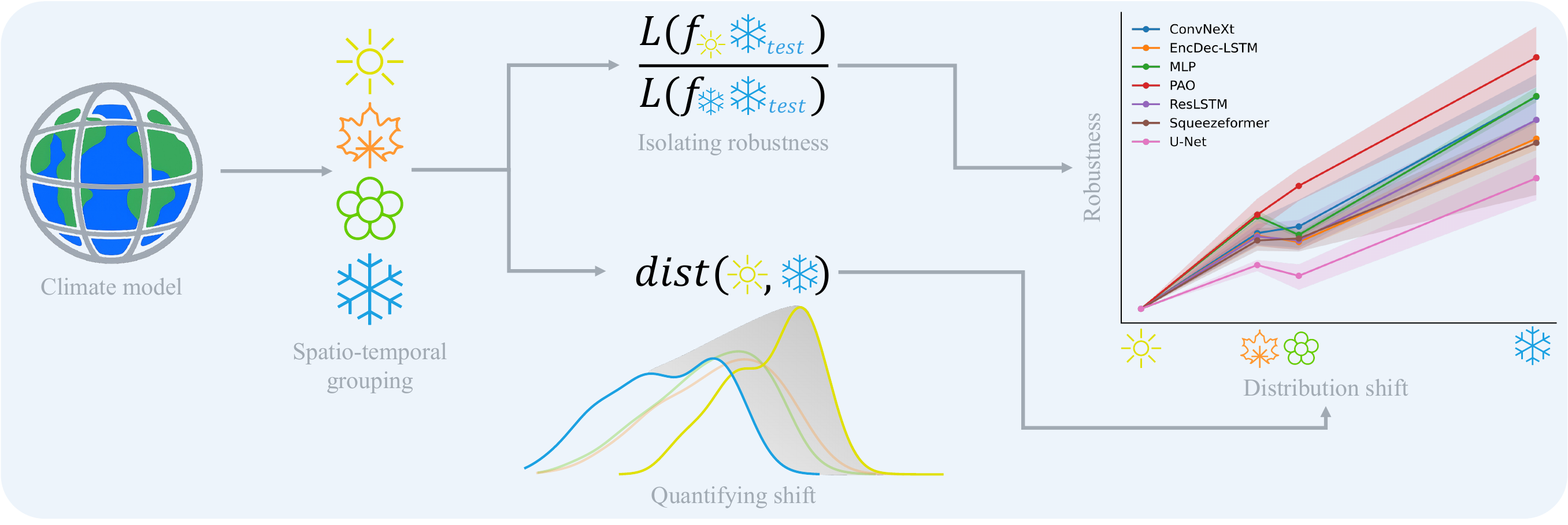}
    \caption{A zero-overhead framework for evaluating emulator robustness. Inspired by emergent constraints in climate science, we demonstrate that seasonal cycles induce structured distribution shifts that serve as a rigorous, real-world proxy for long-term climate change. By partitioning existing data into distinct seasonal regimes (e.g., Northern Hemisphere winter vs. summer), we construct a realistic out-of-distribution testbed without additional data collection or synthetic perturbations. This approach allows us to quantify the \textit{robustness gap} in state-of-the-art hybrid-ML emulators by isolating performance degradation under distribution shift from inherent prediction difficulty.}
    \label{fig/methodology}
\end{figure}

% Stability is solved
ML emulator stability for climate emulation is well developed \cite{yu_climsim-online_2025, brenowitz_interpreting_2020, watt-meyer_ace2_2025, kochkov_neural_2024, lin_navigating_2025},  with recent state-of-the-art emulators demonstrating stable behaviour in stand-alone and hybrid configurations for high-resolution \cite{watt-meyer_ace2_2025} and low-resolution \cite{lin_crowdsourcing_2025} real-geography settings, respectively. Robustness, on the other hand, is a more fundamental challenge. \textit{In many ML applications, distribution shift is an undesirable edge case. In climate modelling, it is the problem itself \cite{gentine_could_2018}.} ML emulators are trained on a specific window of data, yet deployed in an unknown future that statistically departs from it. As a consequence, the robustness of ML emulators remains poorly understood. A large proportion of existing work omits out-of-distribution (OOD) evaluation entirely \cite{brenowitz_spatially_2019, gentine_deep_2021, mooers_assessing_2021, han_moist_2020, behrens_simulating_2025, perezhogin_generalizable_2025, hu_stable_2025, lin_crowdsourcing_2025, yu_climsim-online_2025}, and those that do evaluate it consider only a narrow set of scenarios \cite{watt-meyer_ace2_2025, gregory_floenet_2026, kochkov_neural_2024, ogorman_using_2018, beucler_towards_2020, kuhbacher_towards_2024, han_ensemble_2023, beucler_climate-invariant_2024, rackow_robustness_2024, iglesias-suarez_causally-informed_2024, lin_stress-testing_2024}, often relying on physically unrealistic perturbations such as uniform sea surface temperature changes that fail to capture the spatial heterogeneity of realistic climate shifts. Moreover, it remains unclear whether performance on a single simulation scenario reflects general emulator robustness or performance on that specific case alone, rendering existing evaluation protocols narrow and potentially uninformative for assessing realistic model robustness. 

To bridge this gap between current ML practice and the requirements of climate science, we present a framework that redefines the evaluation and design of robust climate emulators:

\begin{enumerate}
\item We characterise climate emulation as a fundamental \textit{out-of-distribution prediction task} and provide quantitative evidence for this framing through analysis of 40 years of observation-constrained reanalysis data;

\item We provide \textit{empirical evidence that seasonal variation serves as a valid proxy} for the distribution shifts induced by long-term climate change, establishing a realistic physical basis for our evaluation methodology;

\item Based on these seasonal proxies, we introduce a \textit{zero-overhead evaluation framework} and use it to demonstrate that current state-of-the-art hybrid-ML emulators exhibit systematic performance degradation under realistic distribution shifts;

\item Finally, we identify \textit{compositional generalisation} as a structural requirement for robustness, and demonstrate that physically-motivated model decompositions significantly improve OOD robustness.

\end{enumerate}

%-- Related work --%
\section{Related work}
\label{section/related_work}

ML has enabled substantial advances in climate emulation, spanning sub-grid parametrisations that capture specific physical processes (e.g., convection and radiation) \citep{gentine_could_2018, brenowitz_interpreting_2020, lin_navigating_2025, jebeile_machine_2023, perezhogin_generalizable_2025, han_ensemble_2023, behrens_simulating_2025, han_moist_2020, bhouri_multi-fidelity_2023, ogorman_using_2018}, emulators of cloud-resolving models that represent sub-grid dynamics as unified systems \citep{mooers_assessing_2021, lin_crowdsourcing_2025, hu_stable_2025}, and global-scale emulators that model the complete climate system (e.g., ACE2 \citep{watt-meyer_ace2_2025} Neural-GCM \cite{kochkov_neural_2024}, and GenCast \cite{price_probabilistic_2025}). These developments have been supported by an ecosystem of benchmarks and datasets \citep{watson-parris_climatebench_2022, kaltenborn_climateset_2023, rasp_weatherbench_2024, yu_climsim_2023, cachay_climart_2021, nguyen_climatelearn_2023}. We categorise existing work and highlight its relationship to our contributions.

\textbf{Out-of-distribution evaluation in climate modelling.  }
Despite the critical nature of robustness, a vast proportion of research into ML emulators for climate modelling lacks any form of OOD evaluation \cite{brenowitz_spatially_2019, gentine_deep_2021, mooers_assessing_2021, han_moist_2020, behrens_simulating_2025, perezhogin_generalizable_2025, hu_stable_2025, lin_crowdsourcing_2025, yu_climsim-online_2025, nguyen_climax_2023}. Those that do either use highly approximated and physically unrealistic scenarios such as uniform heating and cooling of the sea surface temperature \cite{beucler_towards_2020, kuhbacher_towards_2024, han_ensemble_2023, beucler_climate-invariant_2024, iglesias-suarez_causally-informed_2024, lin_stress-testing_2024}, or utilise more physically grounded scenarios including increasing optical wave thickness \cite{ogorman_using_2018}, long autoregressive rollouts with forcings \cite{kochkov_neural_2024, watt-meyer_ace2_2025}, and use of high-resolution scenario simulations \cite{gregory_floenet_2026, rackow_robustness_2024, cachay_climart_2021, nguyen_climatelearn_2023, navarro_assessing_2026}. However, these approaches remain constrained by their narrow scope, typically involving three or fewer test scenarios, and a reliance on computationally expensive data acquisition. Critically, these methods fail to provide a standardised metric to quantify the magnitude of the distributional shifts being tested, obscuring the relationship between shift severity and model failure. While ClimateBench \citep{watson-parris_climatebench_2022} and ClimateSet \citep{kaltenborn_climateset_2023} represent pivotal benchmarking contributions that partially address OOD evaluation, they primarily frame climate modelling as a supervised mapping from global forcings to annual-mean states. Consequently, these benchmarks prioritize the learning of statistical trends over underlying system dynamics, distinguishing them from the ML climate emulators considered in this study.

\textbf{Characterising model performance under shift.  }
For studies that evaluate generalisability to other climates, the majority report some form of performance degradation \cite{watt-meyer_ace2_2025, kochkov_neural_2024, ogorman_using_2018, beucler_towards_2020, kuhbacher_towards_2024, rackow_robustness_2024, lin_stress-testing_2024}. Performance under distributional shift has been shown to improve through learning climate-invariant transformations \cite{beucler_climate-invariant_2024}, using multiple-model predictions \cite{han_ensemble_2023}, and incorporating physical priors \cite{gregory_floenet_2026}. However, these comparisons implicitly assume that all scenarios are equally predictable. As a result, no existing work accounts for differences in the intrinsic predictability of the underlying dynamics across scenarios. For example, a warmer climate is expected to exhibit more erratic behaviour, with an increased frequency and intensity of extremes \cite{intergovernmental_panel_on_climate_change_ipcc_climate_2023}, making it more difficult to predict. Without accounting for this, there is no fair basis for comparing model performance across different scenarios.

\textbf{Out-of-distribution evaluation beyond climate.  }
In contrast, the broader machine learning community has developed benchmarks for systematically assessing robustness under distributional shift, such as WILDS \citep{koh_wilds_2021} and DomainBed \citep{gulrajani_search_2020}, which emphasise evaluation across multiple, well-defined shift types. Climate emulation studies, by comparison, largely rely on limited scenario-specific evaluations, raising the question of whether current frameworks are sufficient to capture the challenges of generalisation under diverse and realistic shifts.

Together, these limitations, namely (i) limited test scenarios, (ii) computationally expensive data acquisition, and (iii) no standardised metric motivate the development of evaluation frameworks that span a broader and more structured range of distributional shifts, enable systematic and fair comparison of robustness across scenarios, and better reflect real-world deployment conditions.

%-- Experimental Setup --%
% \input{content/experimental_setup.tex}

%-- Climate prediction is OOD prediction --%
\section{Climate prediction is out-of-distribution prediction}
\label{section/climate_prediction_is_ood_prediction}

% *Non-stationarity of climate* 
Anthropogenic forcing is driving the climate system beyond its historical range of variability \cite{intergovernmental_panel_on_climate_change_ipcc_climate_2023}, making non-stationarity an intrinsic property of the prediction problem. In this section, we quantify whether climate change induces a significant OOD shift.

\textbf{Formalising the task.  } We consider a spatio-temporal climate dataset $\mathcal D = \{(x_{t,g}, y_{t,g}) : t=1,...,T, g \in \mathcal G\}$, where each sample is defined on a spatial grid $\mathcal{G} = \{g_1, ..., g_N\}$ (e.g., a latitude-longitude grid). The inputs $x_{t,g} \in \mathbb R^{d_{in}}$ represent the state of the system at time $t$ at location $g$, described by a set of features. The targets $y_{t,g} \in \mathbb R^{d_{out}}$ represent the corresponding state after $\Delta t$ (e.g., 0.5-12 hours). We train an ML emulator to learn a mapping

\begin{equation}
    f_\theta : \mathbb R^{n \times d_{in}} \rightarrow \mathbb R^{n \times d_{out}},  1 \leq n \leq N,
\end{equation}

where $n$ denotes the number of grid cells provided as input. Depending on the modelling objective, an emulator can operate at different spatial scales: from hybrid ML settings, where sub-grid processes are modelled at a single grid cell ($n=1$), to regional emulators over subsets of the domain ($1<n<N$), and up to global emulators that model the full spatial field ($n=N$). In practice, the emulator predicts change in state (tendencies) rather than raw states, which can improve numerical stability.

\textbf{Quantifying shift.  } We adopt model-agnostic metrics from the drift detection literature to measure distributional shift, primarily using energy distance (ED) because of its computational efficiency \citep{goldenberg_survey_2019}. Given a set of datasets \(\{\mathcal D_k\}_{k=1}^K\), let \(X_k = \{x_{t,g} : (t,g) \in \mathcal D_k\}\) denote samples drawn from distribution \(F_k\). The ED between two groups \(i\) and \(j\) is defined as

\begin{equation}
\text{ED}(F_i, F_j)
= 2 \,\mathbb{E}\|X_i - X_j\|
- \mathbb{E}\|X_i - X_i'\|
- \mathbb{E}\|X_j - X_j'\|,
\quad i,j = 1,\dots,K,
\end{equation}

where \(X_k'\) is an independent copy of \(X_k\), and \(\|\cdot\|\) denotes the Euclidean norm. ED defines a metric on distributions that is zero if and only if \(F_i = F_j\), and increases with distributional dissimilarity. It can be interpreted as comparing cross-group distances to within-group variability.

\textbf{Climate change causes shift.  } To quantify climate change shift, we use ERA5 monthly reanalysis data~\cite{hersbach_era5_2020} at 2.5° resolution, spatially averaged across grid points, to construct a multivariate distribution of precipitation, air temperature, 500 hPa geopotential height, and total column precipitable water. The period 1979–2025 is partitioned into five consecutive intervals. We then compute multivariate distances between the baseline period (1979–1988) and each subsequent interval. We assess statistical significance using a two-sample permutation test of the null hypothesis that the samples are drawn from the same distribution. We compare the behaviour of ED with maximum mean discrepancy (MMD) and k-nearest neighbour-based KL divergence. For the latter, dimensionality is reduced using the first two principal components obtained via PCA.

\begin{figure}
    \centering
    \includegraphics[width=1\linewidth]{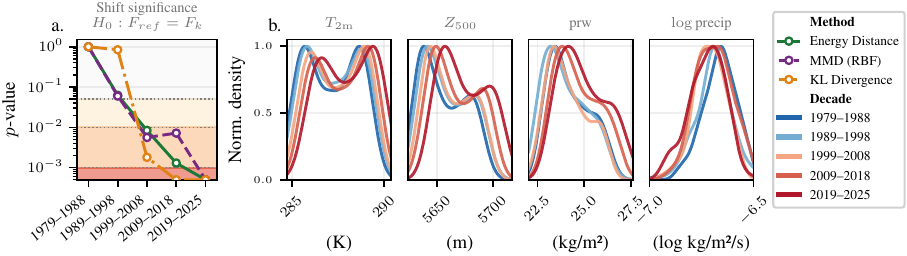}
    \caption{Climate change drives a statistically significant and progressively growing shift in the global atmospheric state distribution. (a.) p-values from two-sample permutation tests comparing each decade to the 1979–1988 baseline under three divergence measures. Decreasing p-values indicate increasing confidence in distributional shift with increasing prediction horizon. (b.) Marginal normalised KDE distributions of global area-weighted monthly means for surface temperature ($T_{2\mathrm{m}}$), 500hPa geopotential height ($Z_{500}$), precipitable water (prw), and log-precipitation, coloured from blue (1979–1988) to red (2019–2025). Systematic shifts are visible in all variables (See Appendix~\ref{appendix/climate_prediction_is_ood_prediction}).}
    \label{fig/climate_shift}
\end{figure}

 Figure~\ref{fig/climate_shift}a reports p-values across metrics and temporal groups. A p-value below $10^{-2}$ provides evidence against the null hypothesis of identical distributions. We observe that p-values decrease with the prediction horizon, implying increasing confidence in the presence of shift. This trend is consistent with expectations: as climate change intensifies over time, the magnitude of distributional shift grows. Figure~\ref{fig/climate_shift}b show marginal distributions for each variable, where shifts across years are qualitatively apparent in all states, supporting the presence of a multivariate OOD regime. Additionally, the consistent trends observed across metrics indicate that the detected shift is not specific to ED, but rather reflects a robust and reliable signal across different measures.

Given that (i) climate emulation inherently models climate change; (ii) we find that climate change induces distribution shift; and (iii) such shifts are known to challenge ML models, we argue that evaluation should place greater emphasis on robustness to distribution shift.

%-- Seasonality as a proxy for climate change -- %
\section{Seasonality as a proxy for climate change}
\label{section/proxy}

% *Why naive solution wont work*
Given the importance of evaluating robustness to distribution shift, a natural starting point is to assess such models using numerical simulations of future climates. However, this strategy is limited by the need for a large ensemble of scenarios to adequately capture the uncertainty inherent in climate models \cite{williamson_emergent_2021}, leading to high computational costs. Even with such ensembles, these simulations may fail to reflect the OOD shifts that the emulator will ultimately encounter. Taken together, these challenges motivate the need for systematic assessment of model behaviour under realistic shifts without relying on explicit future-climate simulations.

\subsection{Seasonal variability as a testbed for out-of-distribution generalisation}
\label{section/climate_prediction_is_ood_prediction/ec_seasonality}

% *Emerging constraints* 
Emergent constraints (ECs) are a class of approaches in climate science that use physically interpretable statistical relationships between observable variability in the present climate and uncertain aspects of future climates to reduce uncertainty in climate projections. The central idea is that if a model fails to reproduce key observable behaviours of the current climate, it is unlikely to reliably simulate the associated future response. Conversely, models that accurately capture these observable relationships can be used to constrain uncertainty in future projections \cite{williamson_emergent_2021}. A large body of EC literature exploits aspects of seasonal variability as the observable constraint \cite{donat_understanding_2018, wenzel_projected_2016, thackeray_emergent_2019, qu_persistent_2014, hall_using_2006, zhai_long-term_2015, knutti_constraining_2006, covey_seasonal_2000} (see Table \ref{tab/seasonal_ec}). For example, \citet{covey_seasonal_2000} and \citet{knutti_constraining_2006} use the seasonal cycle of temperature to constrain the long-term increase in global mean surface temperature following a doubling of atmospheric $\mathrm{CO_2}$, known as the equilibrium climate sensitivity. These statistical relationships provide a measurable, observation-grounded means of assessing whether a model captures the physical relationships that govern future climate responses.

% *How we use this* 
Inspired by this perspective, we propose using seasonal groupings as a testbed for evaluating the robustness of ML emulators. Consistent performance across seasonal regimes indicates that a model has learned meaningful relationships that generalise to future climate conditions. From an ML perspective, seasonal variation can be interpreted as a form of covariate shift. The underlying physical laws governing the climate system—such as conservation of mass, momentum, and energy—remain invariant across seasons as well as under future forcing scenarios \cite{iglesias-suarez_causally-informed_2024}. Seasonal changes therefore alter the distribution of inputs without changing the underlying mapping from state variables to outcomes. In this view, a model that is robust to seasonal variation has learned a representation aligned with the governing rules of the dynamical system. Such a model is therefore more likely to remain reliable under future climate conditions, where similar physical relationships operate under shifted input distributions.

% Physical realness of seasonal shift
Seasonal groupings provide a physically grounded and systematic basis for evaluation, reflecting real transitions in the climate system. They define multiple, related OOD regimes, enabling more diverse assessment of model generalisation across structured shifts in input distributions.

\subsection{Physically interpretable data partitioning}
\label{section/climate_prediction_is_ood_prediction/partitioning}

Restricting the analysis to physically coherent spatial regions reveals physically interpretable patterns of variability. For example, when isolating to Northern Hemisphere mid-latitudes, the transition between winter (DJF)\footnote{DJF: December, January, February; MAM: March, April, May; JJA: June, July, August; SON: September, October, November} and summer (JJA) corresponds to well-understood physical changes, such as increased surface temperatures and reduced snow and ice cover. 

% *How we partition our dataset* 
We construct spatio-temporal groups by partitioning a climate dataset into subsets $\{\mathcal D_k\}^K_{k=1}$, where each group is defined over a spatial domain $\mathcal{G}_k$ and a time window $\mathcal{I}_k$. In this work, we focus on seasonal behaviour across different latitude bands, but groups may be defined using any recurring temporal regimes (e.g., seasons, years, or diurnal cycles) and/or spatial subsets (e.g., latitude bands or regions, land/sea masks). As a concrete example, consider seasonal groupings over the Northern Hemisphere’s mid-latitudes ($\mathcal{G}_{\text{nhmid}} \subseteq \mathcal{G}$). Let $s(t) \in \{\text{DJF}, \text{MAM}, \text{JJA}, \text{SON}\}$ denote the mapping from each time index $t$ to its corresponding meteorological season. We define seasonal groups as

\begin{equation}
    \mathcal D_s = \{(x_{t,g}, y_{t,g}) : s(t) = s, g \in \mathcal G_{nhmid}\},
\end{equation}

\subsection{Seasonal generalisation transfers to climate change regimes}
\label{section/climate_prediction_is_ood_prediction/results_transfers}

Robustness literature has established that in-distribution (ID) performance can be predictive of OOD performance across model families under certain constraints \cite{miller_accuracy_2021, sanyal_accuracy_2025}: better models ID tend to remain better models OOD. We ask whether this structure extends between two distinct OOD regimes, specifically whether OOD performance under a seasonal shift predicts OOD performance under a climate-change shift.

We use ERA5 monthly reanalysis data~\cite{hersbach_era5_2020} at 2.5° resolution to predict log-transformed monthly precipitation at each grid point, from near-surface air temperature, 500hPa geopotential height, and total column precipitable water. Model performance is evaluated across 200 multilayer perceptrons (MLPs) sampled via random search over architecture and optimisation hyperparameters (see Table \ref{tab/mlp_variants}). Each architecture is trained under two independent experimental splits that operate on the same spatial region: (i) seasonal, where models are trained on the cooler season (e.g., DJF in the Northern Hemisphere) across all years in a given region and evaluated on the warmer season (e.g., JJA) under the same constraints; (ii) climate change, where models are trained on all months from 1979–1988 and evaluated on all months from 2019–2025 within the same region (see Appendix~\ref{appendix/climate_prediction_is_ood_prediction}). Models whose training RMSE exceeds the 90th percentile in either split are excluded as failed training runs. All remaining architectures contribute one paired observation, $(\mathrm{RMSE}\mathrm{seas},\ \mathrm{RMSE}_\mathrm{CC})$, reported in original $\log(\mathrm{pr})$ units.

\begin{figure}
    \centering
    \includegraphics[width=1\linewidth]{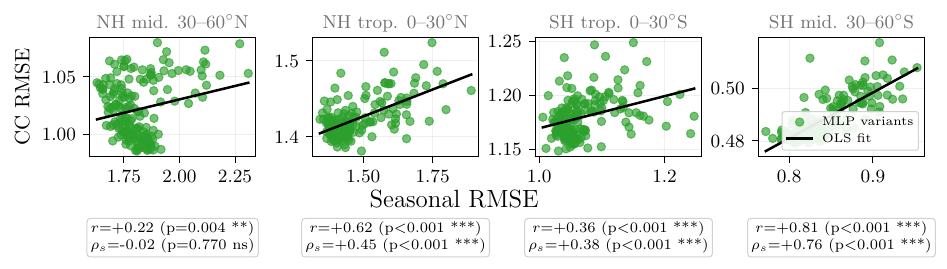}
    \caption{Seasonal OOD performance predicts climate-change OOD performance across regions. Each panel plots seasonal test RMSE against climate-change test RMSE (trained 1979--1988, tested 2019--2025) across up to 200 MLP variants, with Pearson $r$, Spearman $\rho_s$, and an OLS fit overlaid. Positive correlations are observed in all regions, though correlation strength varies. Where climate change produces negligible inter-architecture variation relative to the seasonal shift, correlation estimates become unreliable; the OOD/ID ratio (Figure \ref{fig/seas_proxy_DJF_trained_all}) provides an a priori diagnostic for this.
}
    \label{fig/results_transfers}
\end{figure}

Seasonal OOD performance is a statistically significant predictor of climate-change OOD performance in three of the four regions (Figure \ref{fig/results_transfers}). Architectures that generalise better across seasons also generalise better to a 40-year climate shift, establishing that performance over seasonal shift is an effective proxy for performance over climate change. When the shift caused by climate change is comparably weak relative to the seasonal shift, we find that the correlation estimates become unreliable. Figures~\ref{fig/seas_proxy_JJA_trained_all} and~\ref{fig/seas_proxy_DJF_trained_all} show that when the climate-change perturbation is too weak to differentially affect architectures, the correlation is dominated by sampling noise and its sign is uninformative. The seasonal proxy should therefore be expected to hold only where the climate-change signal is comparatively strong enough to rank models, a condition diagnosable a priori from the OOD/ID ratio.

%-- ML emulators are not robust --%
\section{ML emulators are not robust}
\label{section/ml_emulators_are_not_robust}

Having established seasonal shift as a suitable test-bed for evaluating robustness, we now describe the training and evaluation protocol used to quantify emulator robustness, apply the framework to the ClimSim dataset \cite{yu_climsim-online_2025} and present the resulting empirical findings. Here we focus on ML emulators that predict the large-scale grid tendencies of atmospheric state variables (e.g. wind and moisture) that occur due to the subgrid physics (e.g. convection and clouds), which are driven by large-scale state variables (e.g. temperature, humidity); such emulators are also known as parametrisations.

\textbf{Training.  } For each group \(\mathcal{D}_k\), we construct disjoint training, validation, and test splits. A model is trained on the training subset of a single group, monitored with its validation set, and subsequently evaluated on the test sets of all groups. This procedure is repeated for each group, yielding a collection of models \(\{f_{\theta_k}\}_{k=1}^K\), where \(f_{\theta_k}\) denotes a model trained on \(\mathcal{D}_k\). To reflect a realistic deployment setting, we treat shifted groups as unseen future states. Consequently, normalisation statistics are computed exclusively from the training data.

\textbf{Model evaluation.  } Directly comparing performance between different OOD testsets, e.g., \(L(f_{\theta_k}(X_i))\) versus \(L(f_{\theta_k}(X_j))\), does not account for intrinsic differences in predictability, such as varying levels of variability, noise, or prevalence of extreme events. To control for this, we normalise performance using an ID baseline and define the relative error,
\begin{equation}
    e_r(i,j) = \frac{L(f_{\theta_i}(X_j))}{L(f_{\theta_j}(X_j))}, \quad i,j = 1, \dots, K,
    \label{equation/relative_error}
\end{equation}
where $L$ is the loss function. This ratio isolates the degradation in performance due to distributional shift, independent of the inherent difficulty of the test group. A model that maintains low relative error across increasing distribution shift is considered more robust.

\textbf{ClimSim.  } ClimSim \citep{yu_climsim-online_2025} is a benchmark for ML emulators of sub-grid physical processes derived from a 10-year stationary simulation of the Energy Exascale Earth System Model with Multi-scale Modelling Framework (E3SM-MMF) \cite{hannah_checkerboard_2022,hannah_separating_2021,hannah_initial_2020,norman_unprecedented_2022}. E3SM-MMF couples a host model for large-scale dynamics with an embedded high-resolution, cloud-resolving model (CRM) that serves as the emulation target. While ClimSim has supported progress in accurate and stable emulation through a large-scale Kaggle competition \citep{lin_crowdsourcing_2025, hu_stable_2025}, it has no OOD evaluation. We use this as a case study to showcase how our framework can enable quasi-OOD evaluation via partitioning.

Baseline models include five top-performing Kaggle submissions from \citet{lin_crowdsourcing_2025}, the U-Net of \citet{hu_stable_2025}, and a single-layer MLP as a reference; full details are in Appendix \ref{section/appendix/degrades}.

\subsection{Model performance degrades with distributional shift} We partition ClimSim into seasonal groups (DJF, MAM, JJA, SON) across four regions (Northern and Southern Hemisphere mid-latitudes and tropics), training each model on one season and evaluating on all seasons within the same region across three random seeds. This yields 336 trained models and 1008 OOD test results, from which we compute the relative error $e_r$ for each season-region-model combination, where $L$ is the MAE. As models predict 308 output features, we additionally group these into physically motivated variable groups (Table \ref{table/physical_process_groups}).

\begin{figure}[h]
    \centering
    \includegraphics[width=1\linewidth]{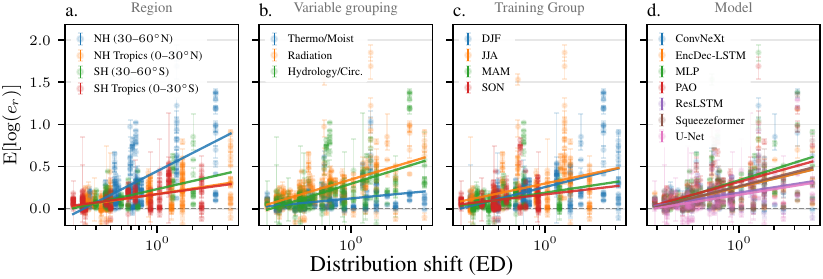}
    \caption{Current hybrid-ML emulators degrade systematically with increasing distribution shift. Each panel shows error ratio against energy distance for 336 trained models, grouped by region, variable group, season, and architecture, respectively.}
    \label{fig/degrades}
\end{figure}

Figure \ref{fig/degrades} plots the expected log relative error $\mathbb{E}[\log(e_r)]$ against multivariate distribution shift measured by ED, with a line of best fit overlaid per group. A perfectly robust performance would have zero slope. We find a systematic positive relationship between distribution shift and performance degradation that holds across regions, variable groups, train/test season combinations, and model architectures (see Appendix~\ref{section/appendix/degrades}). 

This finding has a pointed implication: no current emulator has learned a sufficiently good approximation of the underlying physical dynamics to generalise beyond its training distribution. Yet generalisation to unseen climates is precisely what these emulators are needed for. Our results demonstrate that straightforward partitioning of existing data is sufficient to expose this gap conclusively, and that robustness must be treated as a first-class requirement in both the evaluation and development of ML climate emulators.

We observe apparent breaks in the degradation trend and categorise them as two cases: $e_r$ not increasing with shift, and $e_r < 1$. The first appears to arise because multivariate shift can mask feature-relevant distributional differences for certain output variables (Section~\ref{section/appendix/results/degrades/radiation_vs_lag}); the second suggests that some seasons may provide a better training distribution for certain output features than that feature's own ID data (Section~\ref{section/appendix/results/degrades/er_less_0}).

%-- Composing for OOD robustness --%
\section{Composing for out-of-distribution robustness}
\label{section/composing}

% Big context, 
Compositional generalisation (CompGen) refers to the ability to form novel combinations from previously observed elementary components. Inspired by human language and visual perception \cite{behrens2018cognitive, mitchell2021abstraction, murphy2004big, tenenbaum2011grow}, this concept has been increasingly explored in ML for image and language tasks \cite{fodor_connectionism_1988, goyal2022inductive, greff2020binding, lake2017building}. Here, we investigate whether similar principles can improve the robustness of hybrid ML emulators for climate.

\begin{figure}[t!]
    \centering
    \includegraphics[width=1\linewidth]{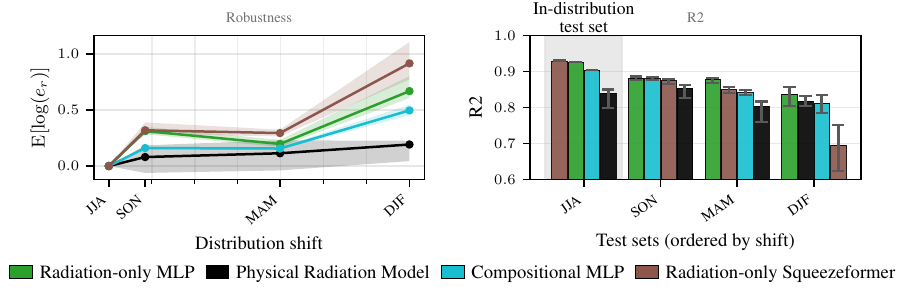}
    \caption{Composing experts improves robustness under covariate shift without sacrificing ID skill. Models trained on JJA to predict surface radiation fluxes are tested on all seasons. Left: Mean log error ratio $\mathbb{E}[\log(e_r)]$ (lower = more robust). Right: $R^2$ on each test set ordered by covariate shift. Shading shows min–max over 10 seeds. The physical model is the most robust but the least skilful ID. Standard ML baselines (MLP, Squeezeformer) are skilful but degrade sharply with shift. The compositional MLP, which learns separate experts per physically grounded regime, improves robustness over all ML baselines while substantially exceeding the physical model's ID $R^2$.}
    \label{fig/composing}
\end{figure}

% Radiation 
We focus on emulating radiation processes in the ClimSim dataset, that is, net shortwave (NETSW) and downward longwave (FLWDS) fluxes at the surface. By first implementing a classical radiation parameterisation, interpretable as a physically informed, underparameterised model, we tackle a central challenge for CompGen in climate modelling: identifying the appropriate elementary components. Building on this insight, we introduce a compositional MLP that aims to balance robustness and predictive skill, while leaving several open challenges. Figure~\ref{fig/composing} summarises the key results, which we discuss below.

\textbf{Physical models are robust.  } The physical model is a piecewise parameterisation separating clear-sky and cloudy-sky regimes. Shortwave fluxes are computed from incoming solar radiation and solar geometry, modulated by effective albedo and cloud transmittance, while longwave fluxes are estimated via a Stefan–Boltzmann formulation with atmospheric emissivity parameterised as a function of humidity, cloud condensate, and pressure. The model is calibrated on the training set using nonlinear least squares; full details are provided in Appendix~\ref{section/appendix/composing}.

% Left panel results and insight
The left panel of Figure~\ref{fig/composing} shows that the physical model is substantially more robust than any ML baseline under seasonal shift. We attribute this to its functional form: 

\begin{enumerate}
    \item \textbf{Regime partitioning: } Splitting the prediction task into physically grounded regimes whose behaviours are consistent across seasons. 
    \item \textbf{Variable selection: } Constraining each regime to operate only on relevant variables.
    \item \textbf{Function description: } Constraining behaviour to the governing equations (e.g. Stefan–Boltzmann) that encode known variable interactions within each regime.
\end{enumerate}

These properties prevent the model from fitting statistical patterns that do not generalise across seasons, anchoring predictions instead in physical relationships that are stable by construction. However, this constrained functional form comes at a cost: with only 38 tuneable parameters, the physical model cannot capture the full complexity of radiation dynamics, and the right panel shows it performs substantially worse than all ML models ID.

\textbf{Data-driven experts bridge the robustness gap.  } Taking inspiration from the physical model, we implement a compositional MLP that isolates the effect of structured decomposition (regime partitioning and variable selection) from other aspects of the physical model; namely its fixed function description, underparameterisation, and constrained optimisation. Separate MLP experts are learnt for each regime and output feature, with only input features deemed relevant to that regime. Crucially, the compositional MLP follows standard ML practice: normalisation, hyperparameter search, and gradient-based training. The prediction is then made as a weighted combination of expert outputs.

Figure~\ref{fig/composing} shows that the compositional MLP improves robustness over all other ML approaches while maintaining higher ID skill than the physical model. We draw two conclusions from this. First, structuring a model around physically meaningful, season-invariant partitions improves OOD generalisation; that is, by learning regime-specific behaviours independently, the model composes more reliably under novel conditions. Second, by relaxing the fixed functional description while preserving the same decomposition structure, the model is free to learn complex radiation dynamics that cannot be captured by simple closed-form expressions, recovering substantial ID skill compared to the physical model.

%-- Current evaluation frameworks are lacking --%
% \input{content/current_evaluation_frameworks_are_lacking}

%-- Conclusion --%
\section{Conclusion}
\label{section/conclusion}

We confirm that climate change causes significant multivariate distribution shifts, demonstrating that robustness to shift is a fundamental requirement for ML climate emulators. We then establish that seasonal variation serves as a structured, zero-overhead proxy for climate-change-induced shift, and use this to show that current state-of-the-art hybrid-ML emulators degrade systematically under realistic distributional shifts, suggesting that the standard ML approach requires rethinking for the climate problem. Finally, we identify compositional generalisation as a principled route forward, demonstrating that learning experts over climate-invariant regimes improves OOD robustness.

Several limitations point towards future work. Our ERA5 analysis is restricted to a reduced variable set, a single prediction task, and monthly timesteps, whilst our ClimSim evaluation is confined to architectures within the existing benchmark (Appendix \ref{section/appendix/limitations}). More fundamentally, our compositional approach currently relies on explicit domain knowledge to define regimes, limiting its scope to what can be specified a priori. Data-driven regime identification offers a path beyond this constraint, shifting from physics-informed to physics-learned decompositions and opening the door to robust emulation without dependence on prescribed physical structure.

%-- Acknoledgments --%
\section{Acknowledgments}
This work was conducted as part of the Intelligent Earth CDT supported by funding from the UK Research and Innovation Council (UKRI) grant number EP/Y030907/1. This work used JASMIN, the UK’s collaborative data analysis environment ( https://www.jasmin.ac.uk) Ingmar Posner holds concurrent appointments as a Professor of Applied AI at the University of Oxford and as an Amazon Scholar. This paper describes work performed at the University of Oxford and is not associated with Amazon. 

Hannah Christensen is supported by a Leverhulme Trust Research Leadership Award `Seamless Uncertainty Quantification for Earth System prediction’ (SUQCES) and through the EERIE project (Grant Agreement No 101081383) funded by the European Union. Views and opinions expressed are however those of the authors only and do not necessarily reflect those of the European Union or the European Climate Infrastructure and Environment Executive Agency (CINEA). Neither the European Union nor the granting authority can be held responsible for them. University of Oxford’s contribution to EERIE is funded by UK Research and Innovation (UKRI) under the UK government’s Horizon Europe funding guarantee (grant number 10049639).

For the purpose of Open Access, the author has applied a CC BY public copyright licence to any Author Accepted Manuscript version arising from this submission.

The authors would also like to thank Anissa Alloula and Markus Baumgartner, whose keen insight and invaluable discussions were instrumental in making this work possible.

%%%%%%%%%%%%%%%%%%%%%%%%%%%%%%%%%%%%%%%%%%%%%%%%%%%%%%%%%%%%

%-- References --%
\bibliographystyle{unsrtnat}
\bibliography{references_tidy}

%-- Appendix --%
\newpage
\appendix

%-- Climate prediction is OOD prediction --%
\section{Climate prediction is out-of-distribution prediction}
\label{appendix/climate_prediction_is_ood_prediction}

Section \ref{section/climate_prediction_is_ood_prediction} investigates how climate change causes OOD shift. Here we provide the details for the experiment.

\subsection{Data and preprocessing}
\label{appendix/climate_prediction_is_ood_prediction/data}

\paragraph{Variables.}
We use ERA5 monthly reanalysis~\citep{hersbach_era5_2020} at 2.5\textdegree{} horizontal resolution, spanning January 1979 to December 2025 (564 months). Four variables are retained: 2\,m air temperature ($T_{2\mathrm{m}}$, K), 500\,hPa geopotential height ($Z_{500}$, m), total column precipitable water (prw, kg\,m$^{-2}$), and precipitation ($\mathrm{pr}$, kg\,m$^{-2}$\,s$^{-1}$). Precipitation is log-transformed prior to all computations, $\tilde{\mathrm{pr}} = \log(\mathrm{pr} + \varepsilon)$ with $\varepsilon = 10^{-8}$\,kg\,m$^{-2}$\,s$^{-1}$, to reduce skewness.

The four variables are chosen to span the principal axes of climate-change-driven variability in the atmospheric state. Surface temperature ($T_{2\mathrm{m}}$) provides the most direct thermodynamic signal of anthropogenic forcing, exhibiting consistent and spatially broad trends that reflect the integrated radiative imbalance. The 500\,hPa geopotential height ($Z_{500}$) captures large-scale dynamical changes: as the troposphere warms, the atmospheric column expands and geopotential heights rise, while concurrent shifts in jet streams and the Hadley circulation alter the spatial structure of mid-tropospheric flow. Precipitable water (prw) encodes the moisture response, which follows the Clausius--Clapeyron relation and increases with warming. Precipitation ($\log\,\mathrm{pr}$) represents the hydrological response: while its global mean change is constrained by the energy budget, regional intensification and redistribution are expected as the warmer atmosphere holds more moisture and large-scale circulation patterns shift. Critically, these variables are not independent, so using them jointly in a multivariate divergence measure captures shifts in their co-structure that would be invisible to marginal comparisons of individual variables. Together they constitute a compact yet physically comprehensive representation of the atmospheric state relevant to climate change.

\paragraph{Normalisation.}
Each variable is standardised using statistics computed over the full period 1979–2025 and all $N_g = 10{,}512$ global grid cells (73 latitudes $\times$ 144 longitudes):
\begin{equation}
    \tilde{x}^{(v)}_{t,g}
    = \frac{x^{(v)}_{t,g} - \mu^{(v)}}{\sigma^{(v)}},
    \qquad
    \mu^{(v)} = \frac{1}{T N_g} \sum_{t,g} x^{(v)}_{t,g},
    \quad
    \sigma^{(v)} = \mathrm{std}_{t,g}\!\left(x^{(v)}_{t,g}\right),
\end{equation}
where the mean and standard deviation are computed as unweighted averages over all (time, grid-cell) pairs, and the superscript $(v)$ indicates that normalisation is performed independently per variable.

\paragraph{Spatial averaging for the significance test.}
To isolate the large-scale climate signal from within-region spatial variability, each monthly field is reduced to a single area-weighted global mean vector $\bar{x}_t \in \mathbb{R}^4$ before computing divergence statistics:
\begin{equation}
    \bar{x}^{(v)}_t = \sum_{g=1}^{N_g} w_g\, \tilde{x}^{(v)}_{t,g},
    \qquad
    w_g = \frac{\cos(\phi_g)}{\sum_{g'} \cos(\phi_{g'})},
\end{equation}
where $\phi_g$ is the latitude of grid cell $g$. This yields one 4-dimensional sample per month: 120 samples for the reference decade (1979–1988) and between 84 and 120 samples for each comparison period.

\paragraph{Marginal distributions.}
The marginal KDE curves in Figure~\ref{fig/climate_shift} are computed from the same area-weighted spatial means, but expressed in original physical units (i.e.\ without the normalisation step above) to facilitate physical interpretation.

\subsection{Hypothesis test formulation}
\label{appendix/climate_prediction_is_ood_prediction/test}

\paragraph{Null hypothesis.}
For each comparison decade $k$, we test
\begin{equation}
    H_0: F_{\mathrm{ref}} = F_k,
\end{equation}
where $F_{\mathrm{ref}}$ is the joint distribution of global monthly means during the reference period (1979–1988) and $F_k$ is the corresponding distribution during decade $k$. Rejection of $H_0$ implies that the multivariate atmospheric state distribution has shifted.

\paragraph{Permutation test.}
Let $\mathbf{X}_{ref} = \{\bar x_t\}_{t \in \mathcal{T}_{\mathrm{ref}}}$ and $\mathbf{X}_{k} = \{\bar y_t\}_{t \in \mathcal{T}_k}$ denote the two sample sets. We proceed as follows:
\begin{enumerate}[leftmargin=*, itemsep=2pt]
    \item Compute the observed test statistic $d = dist(\mathbf{X}_{ref}, \mathbf{X}_k)$ on the actual data.
    \item Pool the samples: $\mathbf{Z} = \mathbf{X}_{ref} \cup \mathbf{X}_k$.
    \item For each permutation $b = 1, \dots, B$: randomly, evenly split $\mathbf{Z}$ to obtain $\mathbf{A}^{(b)}, \mathbf{B}^{(b)}$; compute $d^{(b)} = dist(\mathbf{A}^{(b)}, \mathbf{B}^{(b)})$.
    \item Estimate the p-value as $p = \bigl(\#\{b : {d}^{(b)} \geq d\} + 1\bigr) \,/\, (B+1)$.
\end{enumerate}
Under $H_0$, the label assignment is exchangeable, so the permutation distribution provides a valid reference. We use $B = 1000$ permutations for ED and MMD, and $B = 500$ for KL; the minimum resolvable p-values are therefore approximately $10^{-3}$ and $2 \times 10^{-3}$, respectively. The subsample size $s = \min(|\mathbf{X}_{ref}|, |\mathbf{X}_k|)$ ensures balanced groups.

\subsection{Divergence measures}
\label{appendix/climate_prediction_is_ood_prediction/metrics}

\paragraph{Energy distance (ED).}
The energy distance between distributions $F_{ref}$ and $F_k$ is
\begin{equation}
    \mathrm{ED}(F_{ref}, F_k)
    = 2\,\mathbb{E}\|X_{ref} - X_k\| - \mathbb{E}\|X_{ref} - X_{ref}'\| - \mathbb{E}\|X_k - X_k'\|,
\end{equation}
where $X_{ref}, X_{ref}' \sim F_{ref}$ and $X_k, X_k' \sim F_k$ independently, and $\|\cdot\|$ is the Euclidean norm. ED is non-negative and equals zero if and only if $F_{ref} = F_k$~\citep{goldenberg_survey_2019}. Each expectation is estimated via Monte Carlo by drawing $5 \times 10^5$ random index pairs for the observed statistic and $2 \times 10^5$ pairs per permutation. Computation is performed on GPU using PyTorch.

\paragraph{Maximum mean discrepancy (MMD).}
MMD with a reproducing kernel $k$ is defined as \cite{goldenberg_survey_2019}
\begin{equation}
    \mathrm{MMD}^2(F_{ref}, F_k) = \mathbb{E}[k(X_{ref},X_{ref}')] + \mathbb{E}[k(X_k,X_k')] - 2\,\mathbb{E}[k(X_{ref},X_k)].
\end{equation}
We use the Gaussian (RBF) kernel $k(x, y) = \exp\!\bigl(-\|x-y\|^2 / 2\sigma^2\bigr)$, with bandwidth $\sigma$ set by the median heuristic: $\sigma = \mathrm{median}\bigl\{\|x_i - x_j\| : x_i, x_j \in \mathbf{X}\bigr\}$, estimated from a subsample of 5{,}000 reference-period points. The same GPU Monte Carlo estimator used for ED is applied, replacing the Euclidean norm with the RBF kernel evaluation.

\paragraph{KL divergence (KL).}
KL divergence is not directly estimable in four dimensions from the available sample sizes, so we first reduce dimensionality via PCA fitted on the reference-period samples, retaining the first two principal components (capturing $>99\%$ of variance in the spatial-mean representation). We then apply the $k$-nearest-neighbour estimator of \citet{wang2009divergence}:
\begin{equation}
    \widehat{D_{\mathrm{KL}}}(P \| Q)
    = \frac{d}{n}\sum_{i=1}^{n}\log\frac{\nu_k(x_i)}{\rho_k(x_i)}
    + \log\frac{m}{n-1},
\end{equation}
where $d=2$ is the reduced dimension, $n = |\mathbf{X}_{ref}|$, $m = |\mathbf{X}_k|$, $\rho_k(x_i)$ is the distance from $x_i$ to its $k$-th nearest neighbour within $\mathbf{X}_{ref}$ (excluding itself), and $\nu_k(x_i)$ is the distance from $x_i$ to its $k$-th nearest neighbour in $\mathbf{X}_k$. We use $k = 5$ throughout, implemented via a ball-tree on CPU (scikit-learn). KL divergence is asymmetric; we report $\widehat{D_{\mathrm{KL}}}(F_k \| F_{\mathrm{ref}})$, measuring how much information is lost when approximating the comparison-period distribution with the reference.

\newpage

%-- Seasonality as a proxy for climate change --%
\section{Seasonality as a proxy for climate change}
\label{appendix/proxy}

\subsection{Literature on emergent constraints that utilise seasonal variability}
\label{appendix/proxy/ec_literature}

A substantial body of work in the emergent constraints (EC) literature leverages seasonal variability as an observable proxy for constraining uncertain aspects of future climate projections. These approaches exploit the fact that seasonal variations in the climate system are governed by the same underlying physical processes that drive long-term climate responses. As such, the ability of a model to accurately reproduce seasonal behaviour provides a useful diagnostic of whether it captures physically meaningful relationships.

Table~\ref{tab/seasonal_ec} summarises representative examples of ECs that utilise seasonal variability. Across these studies, observable quantities derived from the seasonal cycle—such as snow-albedo feedback, atmospheric CO$_2$ amplitude, and cloud responses to temperature—are used to constrain a range of future climate sensitivities, including equilibrium climate sensitivity, cryospheric feedbacks, and carbon cycle responses \citep{williamson_emergent_2021}.

A common theme across these works is that seasonal variability provides a structured and information-rich source of variation that can be directly compared to observations. This makes it particularly valuable for evaluating model behaviour, as it offers a physically grounded and measurable test of whether models capture the relationships that govern future climate change.

\begin{table*}[h]
\centering
\small
\caption{Emergent constraints based on seasonal variability. Each constraint relates an observable seasonal behaviour (X) to an uncertain future climate quantity (Y).}
\label{tab/seasonal_ec}
\begin{tabularx}{\textwidth}{p{2.8cm} p{3.2cm} p{3.2cm} X}
\toprule
\textbf{Source} & \textbf{Constrained quantity (Y)} & \textbf{Seasonal observable (X)} & \textbf{Physical idea} \\
\midrule

\citet{hall_using_2006, qu_persistent_2014} 
& Snow-albedo feedback 
& Springtime snow-albedo feedback 
& Seasonal snow retreat modulates albedo in a way that mirrors long-term warming feedbacks \\

\citet{thackeray_emergent_2019} 
& Sea-ice albedo feedback 
& Summer sea-ice albedo feedback 
& Seasonal sea-ice melt processes scale to long-term ice-albedo feedback strength \\

\citet{wenzel_projected_2016} 
& CO$_2$ fertilization 
& Seasonal CO$_2$ amplitude 
& Seasonal carbon uptake reflects sensitivity of the terrestrial carbon cycle to CO$_2$ \\

\citet{covey_seasonal_2000, knutti_constraining_2006} 
& Equilibrium climate sensitivity (ECS) 
& Seasonal temperature cycle 
& Models reproducing seasonal temperature variations better constrain long-term sensitivity \\

\citet{zhai_long-term_2015} 
& Equilibrium climate sensitivity (ECS) 
& Seasonal cloud sensitivity to SST 
& Seasonal cloud responses reflect cloud feedback strength under warming \\

\citet{donat_understanding_2018} 
& Heat extreme frequency 
& Seasonal land atmosphere feedback 
& Seasonal coupling influences amplification of extreme heat events \\

\bottomrule
\end{tabularx}
\end{table*}

\subsection{Data}
\label{appendix/proxy/data}

\paragraph{Source and preprocessing.}
We use ERA5 monthly reanalysis~\cite{hersbach_era5_2020} on a 2.5°$\times$2.5° global grid, covering January 1979 to December 2025 (564 months total). Four variables are extracted: near-surface air temperature (\texttt{tas}), 500\,hPa geopotential height (\texttt{zg500}), total column precipitable water (\texttt{prw}), and monthly-mean precipitation (\texttt{pr}).  Precipitation is log-transformed before all subsequent processing, $y = \log(\mathrm{pr} + \varepsilon)$ with $\varepsilon = 10^{-8}$\,kg\,m$^{-2}$\,s$^{-1}$ ($\approx 0.86$\,mm\,day$^{-1}$), to reduce right-skew.

\paragraph{Prediction task.}
The task is to predict $\log(\mathrm{pr})$ from the three-dimensional input $(\mathrm{tas},\, \mathrm{zg500},\,\mathrm{prw})$ at each spatial gridpoint \emph{independently}.  Each (gridpoint, month) pair constitutes one training or test observation, yielding on the order of $10^5$ samples per split (see Table~\ref{tab/sample_sizes}).

\begin{table}[h]
\centering
\caption{Training and test sample sizes (gridpoints $\times$ months) per region.  
Mid-latitude regions have 1,872 gridpoints; tropical regions have 1,296.}
\label{tab/sample_sizes}
\small
\begin{tabular}{lrrrr}
\toprule
 & \multicolumn{2}{c}{Climate change} & \multicolumn{2}{c}{Seasonal} \\
\cmidrule(lr){2-3}\cmidrule(lr){4-5}
Region & Train & Test & Train & Test \\
\midrule
NH / SH Mid-Latitudes & 202,176 & 157,248 & 237,744 & 263,952 \\
NH / SH Tropics       & 139,968 & 108,864 & 164,592 & 182,736 \\
\bottomrule
\end{tabular}
\end{table}

\paragraph{Regions.}
We evaluate four non-overlapping regions.  For each region, the seasonal split is defined so that training corresponds to the warm or wet season and testing to the contrasting cold or dry season. We explore both training on DJF and testing on JJA, and vice versa.

\begin{table}[h]
\centering
\caption{Regions, latitudinal extent, and seasonal split assignment.}
\label{tab/regions}
\small
\begin{tabular}{llll}
\toprule
Region & Latitude & Train season & Test season \\
\midrule
NH Mid-Latitudes & 30--60°N  & DJF (winter) & JJA (summer) \\
NH Tropics       & 0--20°N   & JJA (wet)    & DJF (dry)    \\
SH Tropics       & 20°S--0°  & DJF (wet)    & JJA (dry)    \\
SH Mid-Latitudes & 30--60°S  & DJF (austral summer) & JJA (austral winter) \\
\bottomrule
\end{tabular}
\end{table}

\paragraph{Experimental splits.}
Each architecture is evaluated under two independent splits:
\begin{itemize}[noitemsep, topsep=3pt]
  \item \textbf{Climate-change (CC):} trained on all months from 1979--1988 ($\approx$108 training months after validation holdout), tested on all months from 2019--2025 (84 months).
  
  \item \textbf{Seasonal:} trained on one season across all years 1979--2025 ($\approx$127 training months after holdout), tested on the contrasting season across all years (141 months).
  
\end{itemize}
Validation is a time-contiguous holdout of the final 10\% of training 
timesteps, used solely for early stopping; it is never used to compute 
reported RMSE values.

\paragraph{Normalisation.}
To prevent data leakage, normalisation statistics are computed \emph{exclusively 
from the training portion} of each split and applied identically to train and 
test data.  For each split $s$ and region $r$, we compute the per-feature mean 
$\boldsymbol{\mu}^{(s,r)}$ and standard deviation $\boldsymbol{\sigma}^{(s,r)}$ 
over all training (gridpoint, month) pairs:
\begin{equation}
  \tilde{\mathbf{x}} = \frac{\mathbf{x} - \boldsymbol{\mu}^{(s,r)}}
                             {\boldsymbol{\sigma}^{(s,r)}}, 
  \qquad
  \tilde{y} = \frac{y - \mu^{(s,r)}_y}{\sigma^{(s,r)}_y}.
\end{equation}

All models receive normalised inputs and produce normalised outputs.  Reported 
RMSE values are \emph{de-normalised} back to original $\log(\mathrm{pr})$ units 
by multiplying residuals by $\sigma^{(s,r)}_y$, ensuring direct comparability 
across splits despite differing training distributions.  Because each split has 
its own normalisation statistics, the absolute RMSE magnitudes differ between 
the CC and seasonal splits; this motivates also reporting the OOD/ID ratio 
(test RMSE divided by training RMSE) as a scale-free measure of generalisation 
degradation.

\subsection{MLP architecture and training}
\label{appendix/proxy/mlp}

\paragraph{Architecture.}
Each model is a fully connected multi-layer perceptron mapping the 
three-dimensional normalised input $(\tilde{\mathrm{tas}},\, 
\tilde{\mathrm{zg500}},\, \tilde{\mathrm{prw}}) \in \mathbb{R}^3$ to a scalar $\hat{\tilde{y}} \in \mathbb{R}$ (normalised log-precipitation).  All hidden layers share the same width; dropout is applied after each hidden activation during training.  The output layer has no activation or dropout.

\paragraph{Hyperparameter search.}
We sample 200 architectures via random search with fixed seed 42, 
independently varying six hyperparameters according to
Table~\ref{tab/mlp_variants}.  The search space is intentionally restricted to 
avoid degenerate training: dropout above 0.25 and weight decay above $10^{-2}$ 
consistently prevented models from fitting the three-input regression within 
the epoch budget.

\begin{table}[h]
\centering
\caption{MLP hyperparameter random-search space.}
\label{tab/mlp_variants}
\small
\begin{tabular}{llll}
\toprule
Hyperparameter & Distribution & Range & Notes \\
\midrule
Number of hidden layers  & Uniform discrete & $\{2,3,4,5,6\}$ & \\
Hidden layer width       & Log-uniform      & $[64,\,1024]$   & Per layer (shared) \\
Dropout rate             & Uniform          & $[0.0,\,0.25]$  & Applied after each activation \\
Weight decay ($\ell_2$)  & Log-uniform      & $[10^{-5},\,10^{-2}]$ & \\
Learning rate            & Log-uniform      & $[5\times10^{-4},\,10^{-2}]$ & Adam optimiser \\
Activation function      & Uniform discrete & \{ReLU, GELU, Tanh\} & \\
\bottomrule
\end{tabular}
\end{table}

\paragraph{Training procedure.}
All models are trained with full-batch gradient descent using the Adam 
optimiser and mean-squared-error loss on normalised targets.  Because the complete training set fits in GPU memory, full-batch training avoids mini-batch noise and 
yields deterministic per-epoch updates.  Training halts at the earlier of 300 epochs or when validation MSE fails to improve by more than $10^{-7}$ for 20 consecutive epochs (early stopping).  At termination the weights achieving 
the lowest validation MSE are restored. 

\paragraph{Quality filter.}
Models whose training RMSE exceeds the 90th percentile of all models in \emph{either} split are excluded from reported correlations as failed training runs (inadequate fitting of the in-distribution task) rather than architecturally meaningful variation.

\paragraph{Statistical tests.}
We quantify the relationship between seasonal and climate-change OOD test RMSE across architectures using two correlation coefficients: the Pearson correlation $r$ and the Spearman rank correlation $\rho_s$.  Both are reported with two-tailed $p$-values.  All correlations are computed on the quality-filtered subset of architectures (training RMSE $\leq$ 90th percentile in both splits; see above), and an OLS best-fit line is overlaid on each scatter panel to visualise the linear trend.

\subsection{Results}
\label{appendix/proxy/results}

\begin{figure}
    \centering
    \includegraphics[width=1\linewidth]{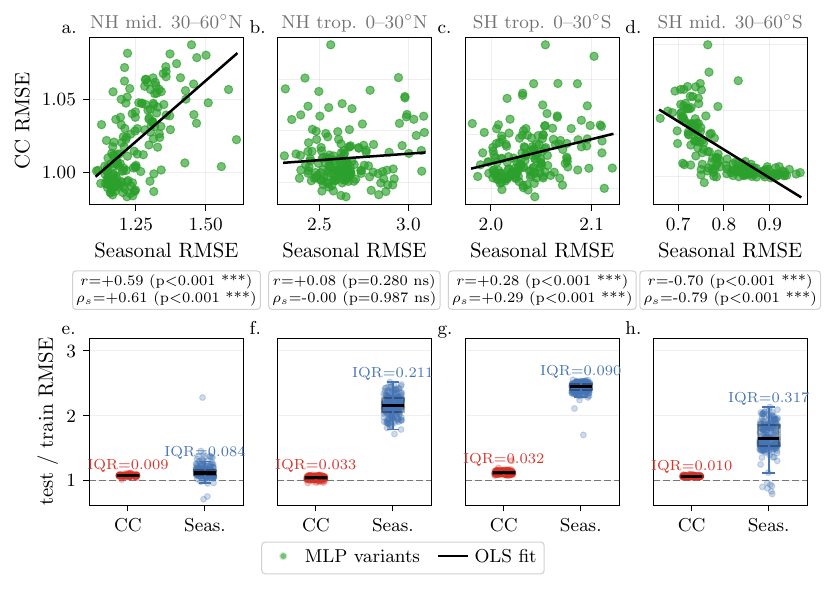}
    \caption{Seasonal OOD performance serves as a useful proxy for climate-change OOD performance when the climate shift signal is sufficiently strong. Panels (a–d) show the relationships, across different regions, between models trained on 1979–1988 and tested on 2019–2025, and models trained on the warmer season and tested on cooler season. Panels (e–h) compare the OOD/ID ratios for each region, demonstrating that the absence of a positive relationship can be anticipated from a comparatively low climate-change OOD/ID interquartile range (IQR).}
    \label{fig/seas_proxy_JJA_trained_all}
\end{figure}

\begin{figure}
    \centering
    \includegraphics[width=1\linewidth]{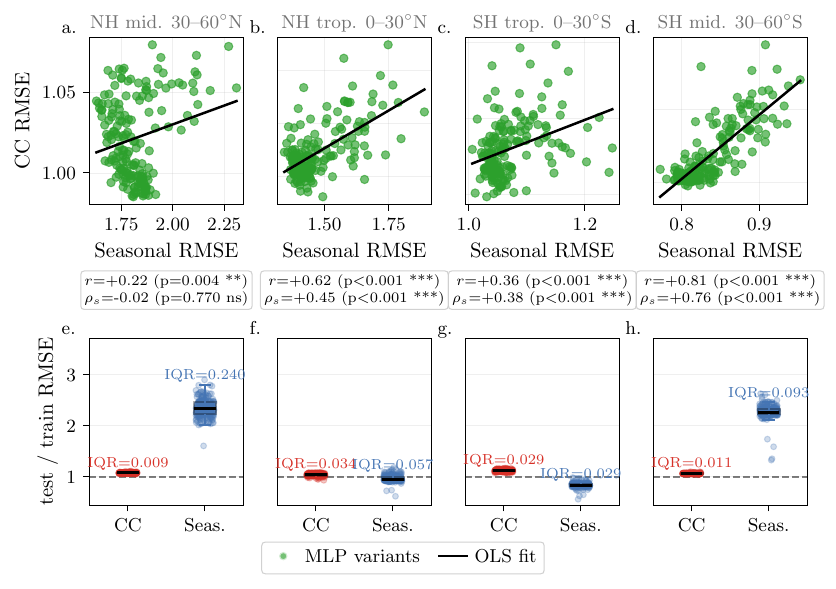}
    \caption{Seasonal OOD performance serves as a useful proxy for climate-change OOD performance when the climate shift signal is sufficiently strong. Panels (a–d) show the relationships, across different regions, between models trained on 1979–1988 and tested on 2019–2025, and models trained on the cooler season and tested on warmer season. Panels (e–h) compare the OOD/ID ratios for each region, demonstrating that the absence of a positive relationship can be anticipated from a comparatively low climate-change OOD/ID interquartile range (IQR).}
    \label{fig/seas_proxy_DJF_trained_all}
\end{figure}

We show results for cool to warm seasons (e.g., DJF to JJA in the Northern Hemisphere and JJA to DJF in the Southern Hemisphere) in Figure~\ref{fig/seas_proxy_JJA_trained_all}, and for warm to cool seasons in Figure~\ref{fig/seas_proxy_DJF_trained_all}.

Seasonal OOD performance predicts climate-change OOD performance where the climate-change signal is strong enough to rank models. Panels (a–d) show that seasonal OOD performance is a statistically significant, positively correlated predictor of climate-change OOD performance when the OOD/ID ratios are sufficiently comparable (e–h). 

\newpage

\newpage

\newpage

%-- ML emulators are not robust --%
\newpage
\section{ML emulators are not robust}
\label{section/appendix/degrades}

\subsection{ClimSim}
\label{section/appendix/degrades/climsim}

ClimSim \citep{yu_climsim-online_2025} is a benchmark for ML emulators of sub-grid physical processes within numerical climate models, derived from a 10-year stationary simulation of the Energy Exascale Earth System Model with Multi-scale Modelling Framework (E3SM-MMF) \citep{hannah_initial_2020, hannah_separating_2021, norman_unprecedented_2022, hannah_checkerboard_2022}. E3SM-MMF couples a low-resolution host model for large-scale dynamics with an embedded cloud-resolving model (CRM) that serves as the emulation target. ClimSim has supported progress in accurate and stable emulation and been widely adopted through a large-scale Kaggle competition \citep{lin_crowdsourcing_2025, hu_stable_2025}. Its evaluation assesses offline predictive skill and online stability, but is restricted to a single stationary regime, leaving behaviour under distributional shift unexplored.

Learning a hybrid-ML emulator can be framed as a high-dimensional, non-linear regression problem. Given input features $x_{t,g} \in \mathbb{R}^{1399}$ representing the local atmospheric state at time $t$ and location $g$, including boundary conditions, vertical structure, and historical large-scale forcing (Table~\ref{table/input_set}), the goal is to predict the state $y_{t,g} \in \mathbb{R}^{308}$ at the same grid point given the subgrid processes (Table~\ref{table/output_set}). The dataset consists of a 10-year rollout of a low-resolution, real-geography simulation with 384 grid points and 60 vertical levels, totalling 100 million samples (744 GB) \cite{yu_climsim-online_2025}. The first 8 years are used for training and validation, with the remaining 2 years held out for testing. We use the input set of \citet{hu_stable_2025}, which includes large-scale memory and achieves state-of-the-art performance; full input and output sets with normalisations are provided in Tables~\ref{table/input_set} and~\ref{table/output_set}.

\begin{table*}
\centering
\caption{List of input features with units and normalisations. 
The table summarizes all atmospheric, surface, and forcing variables used as model inputs along with their respective normalization schemes.}
\label{table/input_set}
\small
\setlength{\tabcolsep}{3pt}
\renewcommand{\arraystretch}{1.15}

\begin{tabular}{@
    {}p{0.23\textwidth}
    @{\hspace{2pt}}
    p{0.07\textwidth}
    p{0.50\textwidth}
    p{0.10\textwidth}
@{}}
\toprule
\textbf{Variable} & \textbf{Units} & \textbf{Description} & \textbf{Norm.} \\
\midrule

\rowstrut $T(z)$ & K & Temperature & $\mathcal{N}_1$ \\
\rowstrut $RH(z)$ &  & Relative humidity & -- \\
\rowstrut $q_n(z)$ & kg/kg & Liquid and ice cloud mixing ratio & $\mathcal{N}_4$ \\
\rowstrut Liquid partition $(z)$ &  & Diagnostic microphysics constraint & -- \\
\rowstrut $u(z)$ & m/s & Zonal wind & $\mathcal{N}_1$ \\
\rowstrut $v(z)$ & m/s & Meridional wind & $\mathcal{N}_1$ \\
\rowstrut $\tfrac{dT_{adv}}{dt}(z,t_0)$ & K/s & Large-scale forcing of temperature at $(t)$ & $\mathcal{N}_2$ \\
\rowstrut $\tfrac{dq_{T,adv}}{dt}(z,t_0)$ & kg/kg/s & Large-scale forcing of total water at $(t)$ & $\mathcal{N}_2$ \\
\rowstrut $\tfrac{du_{adv}}{dt}(z,t_0)$ & m/s$^2$ & Large-scale forcing of zonal wind at $(t)$ & $\mathcal{N}_2$ \\
\rowstrut $\tfrac{dT_{adv}}{dt}(z,t_{-1})$ & K/s & Large-scale forcing of temperature at $(t-1)$ & $\mathcal{N}_2$ \\
\rowstrut $\tfrac{dq_{T,adv}}{dt}(z,t_{-1})$ & kg/kg/s & Large-scale forcing of total water at $(t-1)$ & $\mathcal{N}_2$ \\
\rowstrut $\tfrac{du_{adv}}{dt}(z,t_{-1})$ & m/s$^2$ & Large-scale forcing of zonal wind at $(t-1)$ & $\mathcal{N}_2$ \\
\rowstrut $\tfrac{dT}{dt}(z,t_{-1})$ & K/s & Temperature tendency at $(t-1)$ & $\mathcal{N}_3$ \\
\rowstrut $\tfrac{dq_v}{dt}(z,t_{-1})$ & kg/kg/s & Water vapor tendency at $(t-1)$ & $\mathcal{N}_3$ \\
\rowstrut $\tfrac{dq_n}{dt}(z,t_{-1})$ & kg/kg/s & Total cloud tendency at $(t-1)$ & $\mathcal{N}_3$ \\
\rowstrut $\tfrac{du}{dt}(z,t_{-1})$ & m/s$^2$ & Zonal wind tendency at $(t-1)$ & $\mathcal{N}_3$ \\
\rowstrut $\tfrac{dT}{dt}(z,t_{-2})$ & K/s & Temperature tendency at $(t-2)$ & $\mathcal{N}_3$ \\
\rowstrut $\tfrac{dq_v}{dt}(z,t_{-2})$ & kg/kg/s & Water vapor tendency at $(t-2)$ & $\mathcal{N}_3$ \\
\rowstrut $\tfrac{dq_n}{dt}(z,t_{-2})$ & kg/kg/s & Total cloud tendency at $(t-2)$ & $\mathcal{N}_3$ \\
\rowstrut $\tfrac{du}{dt}(z,t_{-2})$ & m/s$^2$ & Zonal wind tendency at $(t-2)$ & $\mathcal{N}_3$ \\
\rowstrut $O_3(z)$ & mol/mol & Ozone volume mixing ratio & $\mathcal{N}_1$ \\
\rowstrut $CH_4(z)$ & mol/mol & Methane volume mixing ratio & $\mathcal{N}_1$ \\
\rowstrut $N_2O(z)$ & mol/mol & Nitrous volume mixing ratio & $\mathcal{N}_1$ \\
\rowstrut $PS$ & Pa & Surface pressure & $\mathcal{N}_1$ \\
\rowstrut $SOLIN$ & W/m$^2$ & Solar insolation & $\mathcal{N}_2$ \\
\rowstrut $LHFLX$ & W/m$^2$ & Surface latent heat flux & $\mathcal{N}_2$ \\
\rowstrut $SHFLX$ & W/m$^2$ & Surface sensible heat flux & $\mathcal{N}_2$ \\
\rowstrut $TAUX$ & W/m$^2$ & Zonal surface stress & $\mathcal{N}_1$ \\
\rowstrut $TAUY$ & W/m$^2$ & Meridional surface stress & $\mathcal{N}_1$ \\
\rowstrut $COSZRS$ &  & Cosine of solar zenith angle & $\mathcal{N}_1$ \\
\rowstrut $ALDIF$ &  & Albedo for diffuse longwave radiation & $\mathcal{N}_1$ \\
\rowstrut $ALDIR$ &  & Albedo for direct longwave radiation & $\mathcal{N}_1$ \\
\rowstrut $ASDIF$ &  & Albedo for diffuse shortwave radiation & $\mathcal{N}_1$ \\
\rowstrut $ASDIR$ &  & Albedo for direct shortwave radiation & $\mathcal{N}_1$ \\
\rowstrut $LWUP$ & W/m$^2$ & Surface upward longwave flux & $\mathcal{N}_1$ \\
\rowstrut $ICEFRAC$ &  & Sea-ice area fraction & -- \\
\rowstrut $LANtfrac$ &  & Land area fraction & -- \\
\rowstrut $OCNFRAC$ &  & Ocean area fraction & -- \\
\rowstrut $SNOWHICE$ & m & Snow depth over ice & $\mathcal{N}_1$ \\
\rowstrut $SNOWHLAND$ & m & Snow depth over land & $\mathcal{N}_1$ \\
\rowstrut $\cos(\mathrm{lat})$ &  & Cosine of latitude & -- \\
\rowstrut $\sin(\mathrm{lat})$ &  & Sine of latitude & -- \\

\bottomrule
\end{tabular}

\vspace{2pt}
{\footnotesize
\textit{Normalization legend:}
$\mathcal{N}_1(x)=\tfrac{x-\mu}{x_{\max}-x_{\min}}$,
$\mathcal{N}_2(x)=\tfrac{x}{x_{\max}-x_{\min}}$,
$\mathcal{N}_3(x)=\tfrac{x}{\sigma}$,
and $\mathcal{N}_4(x)=1-\exp(-\lambda x)$.
}
\end{table*}
\newpage

\begin{table}[t]
\centering
\caption{List of output variables. $\gamma_1$ and $\gamma_2$ are lower bounds to prevent the output normalizations from creating values that are too large; $\gamma_1 = 3^{-10}$ and $\gamma_2 = 10^{-6}$.}
\label{table/output_set}

\footnotesize
\setlength{\tabcolsep}{3pt}
\renewcommand{\arraystretch}{1.18}
\setlength{\extrarowheight}{2pt}

\begin{tabular}{@
    {}p{0.26\linewidth}
    @{\hspace{2pt}}
    p{0.10\linewidth}
    p{0.44\linewidth}
    p{0.16\linewidth}
@{}}
\toprule
\textbf{Variable} & \textbf{Units} & \textbf{Description} & \textbf{Normalization} \\
\midrule
$\tfrac{dT}{dt}(z,t_0)$ & K/s & Temperature tendency & $x/\mathrm{std}$ \\
$\tfrac{dq_v}{dt}(z,t_0)$ & kg/kg/s & Water vapor tendency & $x/\min(\mathrm{std},\gamma_1)$ \\
$\tfrac{dq_n}{dt}(z,t_0)$ & kg/kg/s & Liquid and ice cloud tendency & $x/\min(\mathrm{std},\gamma_1)$ \\
$\tfrac{du}{dt}(z,t_0)$ & m/s$^2$ & Zonal wind tendency & $x/\min(\mathrm{std},\gamma_2)$ \\
$\tfrac{dv}{dt}(z,t_0)$ & m/s$^2$ & Meridional wind tendency & $x/\min(\mathrm{std},\gamma_2)$ \\
NETSW & W/m$^2$ & Net shortwave flux at surface & $x/\mathrm{std}$ \\
FLWDS & W/m$^2$ & Downward longwave flux at surface & $x/\mathrm{std}$ \\
PRECSC & m/s & Snow rate (liquid water equivalent) & $x/\mathrm{std}$ \\
PRECC & m/s & Rain rate & $x/\mathrm{std}$ \\
SOLS & W/m$^2$ & Downward visible direct solar flux to surface & $x/\mathrm{std}$ \\
SOLL & W/m$^2$ & Downward near-IR direct solar flux to surface & $x/\mathrm{std}$ \\
SOLSD & W/m$^2$ & Downward visible diffuse solar flux to surface & $x/\mathrm{std}$ \\
SOLLD & W/m$^2$ & Downward near-IR diffuse solar flux to surface & $x/\mathrm{std}$ \\
\bottomrule
\end{tabular}
\end{table}
\newpage

\subsubsection{Ensuring a stationary system}
\label{section/appendix/experimental_setup/dataset/prune}

For spatio-temporal grouping, it is important that the held-out test years exhibit similar variability to the training years, so that distributional differences between seasonal groups are consistent across splits. This assumption does not hold in the raw ClimSim dataset. We observe year-to-year drift in the multivariate distribution, primarily originating from the upper atmosphere, attributed to an insufficient one-month spin-up period that leaves the slowly evolving stratosphere out of equilibrium \citep{yu_climsim-online_2025, hu_stable_2025}. Additionally, the top 10 vertical levels of the GCM are unused by the CRM, and the next 5 lie within its sponge layer; these levels contribute noise rather than signal.

\begin{figure}
    \centering
    \includegraphics[width=1\linewidth]{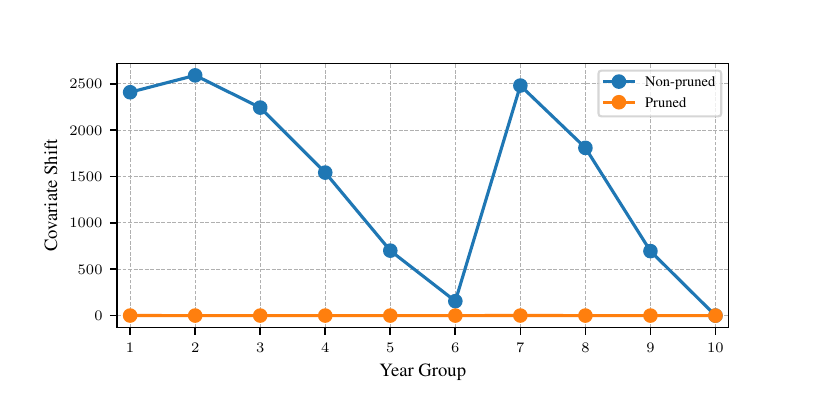}
    \caption{Pruning stratospheric levels enforces stationarity across the dataset. Multivariate energy distance to the final full year is shown for pruned and non-pruned data. Non-pruned data exhibit drift that grows with temporal distance from the reference year, whilst pruned data remain in-distribution throughout. Variables are normalised as $\frac{x - \mu}{\max - \min}$, following \citet{yu_climsim-online_2025}.}
    \label{fig/stationary_plot}
\end{figure}

Following prior work \citep{hu_stable_2025, yu_climsim-online_2025, lin_crowdsourcing_2025}, we set the top 15 vertical levels of all profile variables to zero. Figure~\ref{fig/stationary_plot} confirms that this enforces stationary behaviour across the full dataset.

\subsection{Models}
\label{section/appendix/degrades/models}

We adopt seven baseline architectures: the five top-performing submissions from the ClimSim Kaggle competition \citep{lin_crowdsourcing_2025}, the state-of-the-art UNet of \citet{hu_stable_2025}, and a single-layer MLP as a reference baseline. All models share a unified training configuration (Table~\ref{table/training_config}); full architectural details are provided in \citet{lin_crowdsourcing_2025}.

\paragraph{Squeezeformer.} won the Kaggle competition. It alternates 1D convolution and multi-head self-attention layers to capture local and global dependencies, transforming inputs into $\left(B \times L \times V\right)$ where $B$ is the batch size, $L$ is the number of vertical levels, and $V$ is the number of variables.

\paragraph{Pure ResLSTM.} placed second. It applies a bidirectional LSTM along the vertical dimension with layer normalisation and GELU activation, followed by a linear projection to the output space.

\paragraph{Pao Model.} placed third. It separately processes profile and scalar variables before combining them through a bidirectional LSTM, with variable-type-specific output projections.

\paragraph{ConvNeXt.} placed fourth. Based on \citet{liu_convnet_2022}, it uses a hierarchical structure of depthwise convolutions, normalisation, and channel-wise MLP expansions, following the same input reshaping as Squeezeformer.

\paragraph{Encoder-Decoder LSTM.} placed fifth and has since been adopted in related work \citep{heuer_beyond_2026}. It passes data through an encoder-decoder to produce latent representations before a bidirectional LSTM.

\paragraph{UNet.} is the prior state-of-the-art \citep{hu_stable_2025}, combining a UNet architecture with a self-attention block at the lowest resolution. It achieved stable and skilful online runs over five simulated years.

\paragraph{MLP.} is a single hidden layer network of size 128, included as a reference baseline.

\begin{table}[t]
\centering
\caption{Training hyperparameters used for all models.}
\label{table/training_config}

\small
\setlength{\tabcolsep}{6pt}
\renewcommand{\arraystretch}{1.15}

\begin{tabular}{ll}
\toprule
\textbf{Parameter} & \textbf{Value} \\
\midrule
Batch size & 1024 \\
Epochs & 12 \\
Learning rate & $1\times10^{-4}$ \\
Optimizer & AdamW \\
Loss function & Huber \\
Scheduler & StepLR \\
Step size & 3 \\
Gamma & 0.05 \\
\bottomrule
\end{tabular}
\end{table}

\subsection{Experimental setup}
\label{section/appendix/degrades/setup}

For each of the four regions and four seasonal groups, we train each of the seven architectures with three random seeds, yielding 336 trained models in total. Each model is evaluated on the test sets of all seasons within its training region, and the error ratio $e_r$ is computed for each season-region-model combination, giving 1008 OOD test results. Distribution shift is measured using energy distance computed on the seasonal test sets. The 308 output features are grouped into physically motivated variable groups (Table~\ref{table/physical_process_groups}) to enable process-level analysis.

\begin{table}[h]
\caption{Output variables grouped by physical process. $\gamma_1 = 3^{-10}$ and $\gamma_2 = 10^{-6}$ are lower bounds used in normalization.}
\label{table/physical_process_groups}
\centering
\small
\begin{tabular}{lllll}
\toprule
\textbf{Variable feature groupings} & \textbf{Variables} & \textbf{Description} & \textbf{Units} & \textbf{Normalization} \\
\midrule

\makecell[c]{Thermodynamics \\ \& \\ Moist Processes}
& $\frac{dT}{dt}$ 
& Temperature tendency 
& K s$^{-1}$ 
& $x / \mathrm{std}$ \\

& $\frac{dq_v}{dt}$ 
& Water vapor tendency 
& kg kg$^{-1}$ s$^{-1}$ 
& $x / \min(\mathrm{std}, \gamma_1)$ \\

& $\frac{dq_n}{dt}$ 
& Liquid \& ice cloud tendency 
& kg kg$^{-1}$ s$^{-1}$ 
& $x / \min(\mathrm{std}, \gamma_1)$ \\

\midrule

\makecell[c]{\\ Radiation \\}
& NETSW 
& Net shortwave flux at surface 
& W m$^{-2}$ 
& $x / \mathrm{std}$ \\

& FLWDS 
& Downward longwave flux at surface 
& W m$^{-2}$ 
& $x / \mathrm{std}$ \\

\midrule

\makecell[c]{Hydrology \\ \& \\ Circulation}
& PRECC 
& Rain rate 
& m s$^{-1}$ 
& $x / \mathrm{std}$ \\

& PRECSC 
& Snow rate (liquid water equivalent) 
& m s$^{-1}$ 
& $x / \mathrm{std}$ \\

& $\frac{du}{dt}$ 
& Zonal wind tendency 
& m s$^{-2}$ 
& $x / \min(\mathrm{std}, \gamma_2)$ \\

& $\frac{dv}{dt}$ 
& Meridional wind tendency 
& m s$^{-2}$ 
& $x / \min(\mathrm{std}, \gamma_2)$ \\

\bottomrule
\end{tabular}
\end{table}

\subsection{Results}
\label{section/appendix/degrades/results}

Figure~\ref{fig/degrades} plots $\mathbb{E}[\log(e_r)]$ against energy distance, with results grouped by region, variable group, season, and architecture. The error bars correspond to the min and max mean of the variable groups across the seeds. The loss function, $L$ is MAE. Pearson and Spearman correlations for each grouping are reported in Table~\ref{tab/shift_correlations}; the majority are statistically significant, confirming a systematic positive relationship between distribution shift and performance degradation across all grouping strategies.

\begin{table}[ht]
\centering
\caption{Pearson ($r$) and Spearman ($\rho$) correlations between log error ratio and covariate shift, grouped by analysis dimension. Significance: $^{***}p<0.001$, $^{**}p<0.01$, $^{*}p<0.05$, $^{\text{ns}}p\geq0.05$.}
\label{tab/shift_correlations}
\begin{tabular}{llccccc}
\toprule
\textbf{Dimension} & \textbf{Category} & $r$ & $p_r$ & $\rho$ & $p_\rho$ & $n$ \\
\midrule
  \multirow{4}{*}{\textbf{Region}} & NH ($30$--$60^{\circ}$N) & 0.570 & $^{***}$ & 0.461 & $^{***}$ & 252 \\
   & nh\_tropics\_seasons & 0.454 & $^{***}$ & 0.396 & $^{***}$ & 252 \\
   & SH ($30$--$60^{\circ}$S) & 0.536 & $^{***}$ & 0.258 & $^{**}$ & 150 \\
   & sh\_tropics\_seasons & 0.466 & $^{***}$ & 0.383 & $^{***}$ & 252 \\
\midrule
  \multirow{3}{*}{\textbf{Physical Process}} & Thermo/Moist & 0.321 & $^{***}$ & 0.313 & $^{***}$ & 302 \\
   & Radiation & 0.506 & $^{***}$ & 0.573 & $^{***}$ & 302 \\
   & Hydrology/Circ. & 0.544 & $^{***}$ & 0.587 & $^{***}$ & 302 \\
\midrule
  \multirow{4}{*}{\textbf{Training Group}} & DJF & 0.479 & $^{***}$ & 0.452 & $^{***}$ & 225 \\
   & JJA & 0.351 & $^{***}$ & 0.431 & $^{***}$ & 225 \\
   & MAM & 0.270 & $^{***}$ & 0.336 & $^{***}$ & 228 \\
   & SON & 0.212 & $^{**}$ & 0.108 & $^{\text{ns}}$ & 228 \\
\midrule
  \multirow{7}{*}{\textbf{Model}} & ConvNeXt & 0.465 & $^{***}$ & 0.385 & $^{***}$ & 144 \\
   & EncDec-LSTM & 0.543 & $^{***}$ & 0.484 & $^{***}$ & 144 \\
   & MLP & 0.481 & $^{***}$ & 0.491 & $^{***}$ & 144 \\
   & PAO & 0.475 & $^{***}$ & 0.493 & $^{***}$ & 108 \\
   & ResLSTM & 0.367 & $^{***}$ & 0.370 & $^{***}$ & 114 \\
   & Squeezeformer & 0.525 & $^{***}$ & 0.531 & $^{***}$ & 108 \\
   & U-Net & 0.340 & $^{***}$ & 0.339 & $^{***}$ & 144 \\
\bottomrule
\end{tabular}
\end{table}

\subsection{Breaks from the trend}
\label{section/appendix/results/degrades/trend_breaks}

\subsubsection{$e_r$ does not grow monotonically with shift}
\label{section/appendix/results/degrades/radiation_vs_lag}

For radiation processes, models trained on DJF and JJA generalise better to MAM than to SON, despite MAM being farther from both training distributions in multivariate state space. This apparent contradiction arises because dominant modes of variability in the joint state can mask smaller but more predictive differences in the input--output mapping.

\begin{figure}[h]
    \centering
    \includegraphics[width=0.75\linewidth]{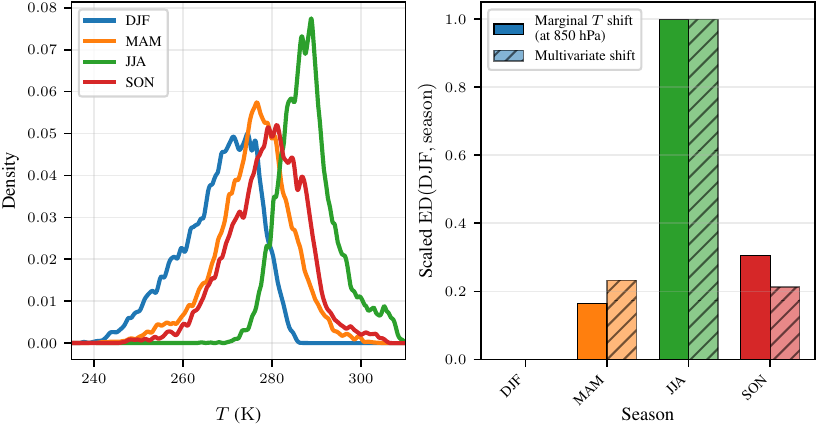}
    \caption{Multivariate distribution shift can obscure similarities in task-relevant variables. Left: seasonal distributions of air temperature at 850hPa, showing MAM is closer to DJF than SON in the thermodynamic state relevant to FLWDS. Right: energy distance to DJF for the full multivariate state (hatched) and for $T$ at 850hPa (solid). MAM is closer to DJF in key radiatively relevant variables despite being farther in the full state space.}
    \label{fig/850hPa_t_vs_multivariate}
\end{figure}

For DJF-trained models, the deviation is driven primarily by predicting downward longwave flux at the surface (FLWDS). FLWDS depends primarily on atmospheric temperature and humidity profiles, and MAM is more similar to DJF in these controlling variables than SON is (Figure~\ref{fig/850hPa_t_vs_multivariate}), consistent with the persistence of cold and dry conditions into early spring and similar cloud--radiation coupling regimes. SON, by contrast, retains residual heat and moisture from summer, shifting the thermodynamic state and thus the mapping to FLWDS.

\begin{figure}[h]
    \centering
    \includegraphics[width=0.75\linewidth]{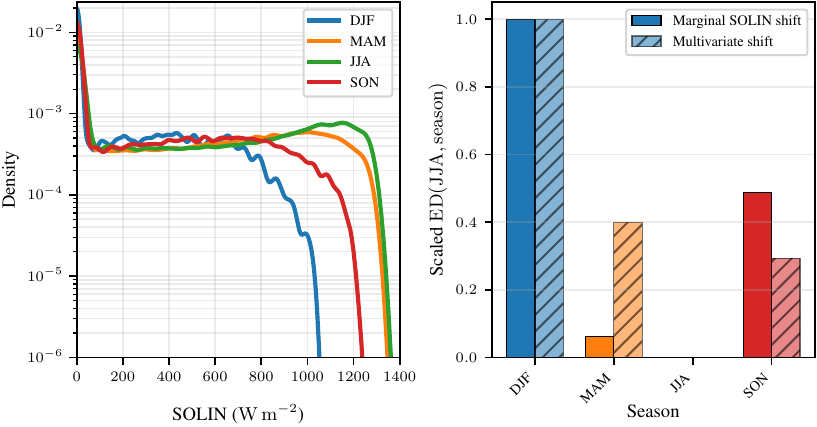}
    \caption{Multivariate distribution shift can obscure similarities in radiatively relevant variables. Left: seasonal distributions of incoming solar radiation (SOLIN), showing MAM is closer to JJA than SON due to the progression of solar forcing. Right: energy distance to JJA for the full multivariate state (hatched) and for SOLIN (solid). MAM is closer to JJA in SOLIN despite being farther in the full state space.}
    \label{fig/SOLIN_vs_multivariate}
\end{figure}

For JJA-trained models, the deviation is driven by net shortwave flux at the surface (NETSW), which is strongly controlled by incoming solar radiation and surface albedo. MAM is more similar to JJA in SOLIN due to increasing insolation (Figure~\ref{fig/SOLIN_vs_multivariate}), whereas SON is characterised by declining insolation and different surface and cloud regimes.

\subsubsection{$e_r < 1$}
\label{section/appendix/results/degrades/er_less_0}

\begin{figure}[h]
    \centering
    \includegraphics[width=0.5\linewidth]{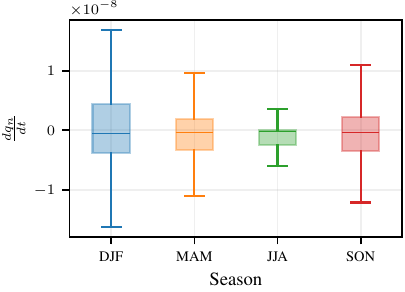}
    \caption{Seasonal distributions of lower-tropospheric (850 hPa) liquid and ice cloud tendency ($\frac{dq_n}{dt}$). Boxes indicate the interquartile range, centre lines the median, and whiskers the non-outlier spread. The JJA distribution is predominantly encapsulated by SON, making SON an effective training distribution for JJA prediction of $\frac{dq_n}{dt}$.}
    \label{fig/liquid_and_ice_cloud_tendency_seasonal_distributions}
\end{figure}

In some cases, models perform better on an OOD test set than on their in-distribution counterpart, yielding $e_r < 1$. This behaviour is primarily isolated to prediction of tropospheric liquid and ice cloud tendency (Figure~\ref{fig/liquid_and_ice_cloud_tendency_seasonal_distributions}). The output distribution for JJA is predominantly encapsulated by that of SON, such that an SON-trained model naturally performs well on JJA targets, which are well-centred around the SON training mean with fewer extremes. In effect, SON provides a better training distribution for predicting JJA cloud tendency than JJA itself. The small magnitude of cloud tendency values is likely attributable to low-resolution grid smoothing.

\newpage

%-- Composing for OOD robustness --%
\section{Composing for out-of-distribution robustness}
\label{section/appendix/composing}

\subsection{Data}
We use the same ClimSim preprocessing and experimental setup as described in Section~\ref{section/appendix/degrades/climsim}. 

\subsection{Problem setup}
We now only look at emulating radiation processes in the ClimSim dataset, that is, net shortwave (NETSW) and downward longwave (FLWDS) fluxes at the surface, and therefore reduce the input features to near-surface temperature, near-surface relative humidity, near-surface liquid and ice cloud mixing ratio, surface pressure (PS), solar insolation (SOLIN), cosine of solar zenith angle (COSZRS), albedo for diffuse shortwave radiation (ASDIF), albedo for direct shortwave radiation (ASDIR), surface upward longwave flux (LWUP), sea-ice area fraction (ICEFRAC), land area fraction (LANDFRAC) and ocean area fraction (OCNFRAC).

\subsection{Models}
We compare four models. The first two are standard ML baselines.

\textbf{MLP.} A fully connected network with 3 hidden layers of 2{,}048 units each, GELU activations, and no dropout. The input is the 12-dimensional radiation feature vector and the output is a 2-dimensional vector predicting NETSW and FLWDS simultaneously. This gives 8{,}423{,}426 trainable parameters.

\textbf{Squeezeformer.} Based on the architecture of \citet{lin_crowdsourcing_2025}, adapted for scalar radiation features. Each of the 12 input features is treated as a token: a learnable feature embedding of dimension 256 is combined with a linear projection of the scalar value, then normalised. This token sequence is passed through 4 Squeezeformer blocks, each comprising a depthwise Conv1D (kernel size 15), multi-head self-attention (4 heads, embedding dimension 256), and a position-wise feed-forward network (expansion factor 4, i.e.\ $256 \to 1024 \to 256$), with residual connections and layer normalisation throughout. The sequence is then mean-pooled and passed through a two-layer head ($256 \to 2048 \to 2$). The total parameter count is 7{,}629{,}314, reduced from the original architecture to match the scale of the radiation-only task.

\textbf{Physical model.} As described in \S\ref{section/appendix/composing/models/physical_model}.

\textbf{Compositional MLP.} Comprised of four expert MLPs — one per (regime, target) combination — each with 3 hidden layers of 512 units, GELU activations, and dropout 0.1. The regimes and their input feature subsets follow the physical model directly: the clear-sky shortwave expert receives 9 solar and surface features (SOLIN, COSZRS, RH, ICEFRAC, LANDFRAC, OCNFRAC, ASDIR, ASDIF, PS), the cloudy shortwave expert receives the same plus cloud condensate qn (10 features), the clear-sky longwave expert receives thermal and moisture features (RH, T, LWUP, PS; 4 features), and the cloudy longwave expert additionally receives qn (5 features), giving expert parameter counts of 530{,}945, 531{,}457, 528{,}385, and 528{,}897, respectively. Each expert predicts a single scalar. The regime weights — a continuous cloud fraction $\alpha \in [0,1]$ and a binary day/night indicator — are computed using the calibrated gating function of the physical model and frozen before expert training begins. Predictions are combined as:
$$\hat{y}\text{NETSW} = d ,\bigl[(1-\alpha),f\text{sw,clear}(\mathbf{x}) + \alpha,f_\text{sw,cloud}(\mathbf{x})\bigr] + (1-d),c_\text{night}$$
$$\hat{y}\text{FLWDS} = (1-\alpha),f\text{lw,clear}(\mathbf{x}) + \alpha,f_\text{lw,cloud}(\mathbf{x})$$
where $d \in \{0,1\}$ is the day/night indicator and $c_\text{night} = -\mu_\text{NETSW}/\sigma_\text{NETSW}$ is the standardised value of zero net shortwave flux. The model has 2{,}119{,}684 trainable parameters. 

All models are trained with AdamW, Huber loss, and a step learning rate schedule; hyperparameters are selected by grid search on a held-out validation split from the training group.

\subsubsection{Physical model}
\label{section/appendix/composing/models/physical_model}

To provide a classical numerical baseline, we implement a piecewise physical parameterisation for surface radiation. The model predicts net shortwave radiation (NETSW) and downward longwave radiation (FLWDS) using physically motivated formulas with a small number of tunable coefficients. The parameterisation is designed to encode key physical mechanisms such as solar geometry, cloud attenuation, and atmospheric emissivity.

The model follows a piecewise structure, separating clear-sky and cloudy-sky regimes via a smooth weighting function, and combining their contributions to produce the final flux estimates. For descriptions of variables below, please refer to Tables~\ref{table/input_set} and~\ref{table/output_set}.

\paragraph{Regime definition (cloud weighting).}
We define a soft weighting to represent the degree of cloudiness in the atmospheric column. First, we define the cosine of the solar zenith angle:
\begin{align}
    \mu_0 &= \max(0, COSZRS)
\end{align}

We then construct a simple cloud proxy:
\begin{align}
    z_{\text{cloud}} &= w_{qn} q_n + w_{rh} RH + w_{sun} (1 - \mu_0)
\end{align}

and map this to a smooth cloudiness weight:
\begin{align}
    w_{\text{cloud}} &= \sigma\big( s \cdot (z_{\text{cloud}} - \tau) \big)
\end{align}

where $\sigma(\cdot)$ is the sigmoid function, $s$ controls the sharpness of the transition, and $\tau$ is a threshold parameter. The weighting satisfies:
\begin{itemize}
    \item $w_{\text{cloud}} \approx 0$: clear-sky regime
    \item $w_{\text{cloud}} \approx 1$: cloudy-sky regime
\end{itemize}

This formulation can be interpreted as a soft gating function that blends between two physically motivated regimes.

\paragraph{Shortwave parameterisation (NETSW).}
NETSW represents the net shortwave radiation absorbed at the surface. We model this as a mixture of clear-sky and cloudy-sky contributions:
\begin{align}
    NETSW = (1 - w_{\text{cloud}}) \cdot NETSW_{\text{clear}} + w_{\text{cloud}} \cdot NETSW_{\text{cloud}}
\end{align}

Both branches share a common physical structure:
\begin{align}
    NETSW_* = SOLIN \cdot \mu_0^{\gamma} \cdot T_* \cdot (1 - \alpha_*)
\end{align}

where $*$ denotes either the clear or cloudy branch, $SOLIN$ is the incoming solar radiation, $\mu_0$ accounts for solar geometry, $T_*$ is atmospheric transmittance, and $\alpha_*$ is an effective albedo. Nighttime conditions are handled implicitly via $\mu_0 = 0$.

\textbf{Clear-sky transmittance:}
\begin{align}
    T_{\text{clear}} = \exp(-k_0 - k_1 RH - k_2 \log(PS))
\end{align}

\textbf{Cloudy-sky transmittance:}
\begin{align}
    T_{\text{cloud}} = \exp(-m_0 - m_1 q_n^{p} - m_2 RH)
\end{align}

\textbf{Surface and effective albedo:}
\begin{align}
    \alpha_{\text{surf}} &= a_0 + a_1 ICEFRAC + a_2 LANDFRAC + a_3 OCNFRAC \\
    \alpha_{\text{cloud}} &= \alpha_{\text{surf}} + a_4 ASDIR + a_5 ASDIF
\end{align}

All albedo terms are constrained to $[0,1]$, and transmittance terms are constrained to $(0,1]$.

\paragraph{Longwave parameterisation (FLWDS).}
FLWDS represents downward longwave radiation at the surface. We again use a mixture formulation:
\begin{align}
    FLWDS = (1 - w_{\text{cloud}}) \cdot FLWDS_{\text{clear}} + w_{\text{cloud}} \cdot FLWDS_{\text{cloud}}
\end{align}

We define a surface temperature proxy using the Stefan--Boltzmann law:
\begin{align}
    T_s = \left(\frac{LWUP}{\sigma_{SB}}\right)^{1/4}
\end{align}

where $\sigma_{SB}$ is the Stefan–Boltzmann constant.

\textbf{Clear-sky longwave:}
\begin{align}
    FLWDS_{\text{clear}} = \epsilon_{\text{clear}} \sigma_{SB} T_s^4 + (1 - \epsilon_{\text{clear}})\sigma_{SB} T^4
\end{align}

\textbf{Cloudy-sky longwave:}
\begin{align}
    FLWDS_{\text{cloud}} = \epsilon_{\text{cloud}} \sigma_{SB} T_s^4
\end{align}

\textbf{Atmospheric emissivity:}
\begin{align}
    \epsilon_{\text{clear}} &= 1 - \exp(-b_0 - b_1 RH - b_2 \log(PS)) \\
    \epsilon_{\text{cloud}} &= 1 - \exp(-c_0 - c_1 RH - c_2 q_n)
\end{align}

Emissivities are constrained to the interval $[0,1]$.

\paragraph{Constraints and numerical stability.}
To ensure physical consistency, we enforce:
\begin{itemize}
    \item $NETSW \geq 0$, $FLWDS \geq 0$
    \item $0 \leq \alpha \leq 1$
    \item $0 \leq \epsilon \leq 1$
    \item $0 < T \leq 1$
\end{itemize}

Clipping or smooth transformations are used where necessary to maintain these bounds.

\paragraph{Parameter estimation.}
All coefficients are calibrated on the training set by minimising mean squared error over both targets:
\begin{align}
    \mathcal{L} = \text{MSE}(NETSW) + \text{MSE}(FLWDS)
\end{align}

Optimisation is performed using a bounded nonlinear optimiser (e.g., L-BFGS-B), with coefficients initialised to physically reasonable values. To improve stability, fitting is performed in stages: clear-sky parameters are first estimated, followed by cloudy-sky parameters, and finally a joint refinement of all coefficients.

The full set of optimised parameters is summarised in Table~\ref{tab/physical_model_parameters}.

\begin{table}[h]
\centering
\caption{Set of tunable parameters in the physical radiation model along with their initial values. Initial values were selected based on physically plausibility, but not tuned.}
\label{tab/physical_model_parameters}
\begin{tabular}{llll}
\toprule
\textbf{Component} & \textbf{Parameter} & \textbf{Description} & \textbf{Initial value} \\
\midrule

\multirow{6}{*}{Regime definition} 
& $c_{\text{sun}}$ & Solar threshold for low-sun regime & 0.18 \\
& $w_{qn}$ & Weight for cloud condensate ($q_n$) & 8.0 \\
& $w_{rh}$ & Weight for relative humidity ($RH$) & 1.4 \\
& $w_{sun}$ & Weight for solar geometry $(1-\mu_0)$ & 0.6 \\
& $\tau$ & Cloud threshold parameter & 0.75 \\
& $s$ & Sigmoid sharpness parameter & 10.0 \\

\midrule

\multirow{5}{*}{Shortwave (clear-sky)} 
& $\gamma$ & Solar zenith scaling exponent & 1.0 \\
& $k_0$ & Baseline optical depth & 0.12 \\
& $k_1$ & Humidity contribution & 0.20 \\
& $k_2$ & Pressure contribution & 0.08 \\
& $k_3$ & Ice-related attenuation & 0.15 \\

\midrule

\multirow{6}{*}{Shortwave (cloudy-sky)} 
& $m_0$ & Baseline cloud optical depth & 0.22 \\
& $m_1$ & Cloud condensate contribution & 6.0 \\
& $m_2$ & Humidity contribution & 0.70 \\
& $m_3$ & Additional cloud proxy contribution & 0.65 \\
& $p_1$ & Exponent for $q_n$ & 0.6 \\
& $p_2$ & Exponent for $RH$ & 1.2 \\

\midrule

\multirow{7}{*}{Albedo} 
& $a_0$ & Base surface albedo & 0.14 \\
& $a_{\text{ice}}$ & Ice fraction contribution & 0.36 \\
& $a_{\text{land}}$ & Land fraction contribution & 0.06 \\
& $a_{\text{ocean}}$ & Ocean fraction contribution & 0.03 \\
& $a_{\text{dir}}$ & Direct radiation contribution (ASDIR) & 0.02 \\
& $a_{\text{dif}}$ & Diffuse radiation contribution (ASDIF) & 0.02 \\
& $a_{\text{low-sun}}$ & Low solar angle correction & 0.08 \\
& $a_{\text{cloud}}$ & Cloud albedo adjustment & 0.07 \\

\midrule

\multirow{3}{*}{Longwave (clear-sky)} 
& $a_0^{lw}$ & Baseline emissivity term & 0.20 \\
& $a_1^{lw}$ & Humidity contribution & 0.90 \\
& $a_2^{lw}$ & Pressure contribution & 0.08 \\

\midrule

\multirow{4}{*}{Longwave (cloudy-sky)} 
& $b_0^{lw}$ & Baseline cloudy emissivity & 0.35 \\
& $b_1^{lw}$ & Humidity contribution & 1.20 \\
& $b_2^{lw}$ & Cloud condensate contribution & 4.50 \\
& $b_3^{lw}$ & Cloud proxy contribution & 0.70 \\

\midrule

\multirow{6}{*}{Temperature proxies} 
& $\delta T$ & Surface-air temperature offset & -1.0 \\
& $t_{rh}$ & RH contribution to air temperature & 2.0 \\
& $t_{qn}$ & Cloud contribution to air temperature & -15.0 \\
& $\Gamma$ & Lapse-rate adjustment for cloud temperature & 8.0 \\
& $t_{rh}^{c}$ & RH contribution to cloud temperature & 1.2 \\
& $t_{qn}^{c}$ & Cloud contribution to cloud temperature & 20.0 \\

\bottomrule
\end{tabular}
\end{table}

\subsection{Hyperparameter search}
\label{section/appendix/composing/hyperparameter_search}

All models share the same training protocol: AdamW optimiser, Huber loss, batch size 1{,}024, a maximum of 12 epochs with early stopping (patience 3, monitoring validation loss), and a step learning rate schedule with step size 3 and decay factor $\gamma = 0.05$. The model checkpoint with the lowest validation loss is selected. Model-specific hyperparameters are tuned by grid search over a single training group (JJA) and a single seed; the selected configuration is then used to train 10 seeds across all four seasonal groups.

Table~\ref{tab/radiation_only_hyperparams} reports the search space and chosen values for each model. The MLP and Squeezeformer sweeps comprised 24 and 48 configurations respectively; the compositional MLP sweep comprised 24 configurations.

\begin{table}[h]
\centering
\caption{Hyperparameter search spaces and chosen values for each radiation-only model. Braces denote the grid searched; the chosen value is \textbf{bold}.}
\label{tab/radiation_only_hyperparams}
\small
\begin{tabular}{llccc}
\toprule
\textbf{Hyperparameter} & \textbf{Search space} & \textbf{MLP} & \textbf{Squeezeformer} & \textbf{Comp.\ MLP} \\
\midrule
\multicolumn{5}{l}{\textit{Architecture}} \\
Hidden / embed dim      & {256, 512, 1024, 2048} & \textbf{2048} & \textbf{256}  & \textbf{512}  \\
Hidden layers (experts) & {1, 2, 3}               & \textbf{3}    & ---           & \textbf{3}    \\
Squeezeformer blocks    & {4, 8, 12}              & ---           & \textbf{4}    & ---           \\
Head dim                & {1024, 2048}            & ---           & \textbf{2048} & ---           \\
Dropout                 & {0.0, 0.1}              & \textbf{0.0}  & \textbf{0.0}  & \textbf{0.1}  \\
\midrule
\multicolumn{5}{l}{\textit{Optimisation}} \\
Learning rate           & {1\text{e-}4,\ 3\text{e-}4} & \textbf{3e-4} & \textbf{3e-4} & \textbf{3e-4} \\
Batch size              & ---                       & 1024          & 1024          & 1024          \\
Optimiser               & ---                       & AdamW         & AdamW         & AdamW         \\
Loss                    & ---                       & Huber         & Huber         & Huber         \\
Max epochs              & ---                       & 12            & 12            & 12            \\
Early stopping patience & ---                       & 3             & 3             & 3             \\
LR step size / $\gamma$ & ---                       & 3 / 0.05      & 3 / 0.05      & 3 / 0.05      \\
\midrule
\multicolumn{5}{l}{\textit{Resulting parameter count}} \\
Parameters              & ---                       & 8{,}423{,}426 & 7{,}629{,}314 & 2{,}119{,}684 \\
\bottomrule
\end{tabular}
\end{table}

\subsection{Training}
\label{section/appendix/composing/training}

Each model is trained independently on each of the four seasonal training groups (JJA, DJF, SON, MAM) of the NH mid-latitude seasons grouping. The train, validation, and test splits are pre-computed and fixed across all runs; their sizes are reported in Table~\ref{tab/radiation_only_splits}. Input features are reduced to near-surface scalars (the deepest vertical level for profile variables), yielding the 12-dimensional feature vector. Feature and target standardisers (zero mean, unit variance) are fit exclusively on the training split of the corresponding group and applied consistently to validation and test data.

To quantify variability due to random initialisation and mini-batch ordering, each model–group combination is trained with 10 random seeds (seeds 0–9), giving 120 training runs per model. The seed controls model weight initialisation and the order in which training batches are sampled; the data split itself is deterministic and does not vary with seed. The checkpoint achieving the lowest validation loss across all epochs is retained for evaluation.

For the compositional MLP, the frozen cloud gate is derived from the physical model calibrated on the same training group, so each group uses its own group-matched gate coefficients. All other models have no dependence on the physical model.

\begin{table}[h]
\centering
\caption{Fixed train / validation / test split sizes per seasonal training group (NH mid-latitude seasons, feature set v6).}
\label{tab/radiation_only_splits}
\small
\begin{tabular}{lccc}
\toprule
\textbf{Group} & \textbf{Train} & \textbf{Validation} & \textbf{Test} \\
\midrule
JJA & 334,152 & 47,736 & 95,328 \\
DJF & 310,536 & 62,712 & 95,328 \\
SON & 330,120 & 47,160 & 95,328 \\
MAM & 333,648 & 47,664 & 95,328 \\
\bottomrule
\end{tabular}
\end{table}

\subsection{Results}
\begin{figure}
    \centering
    \includegraphics[width=1\linewidth]{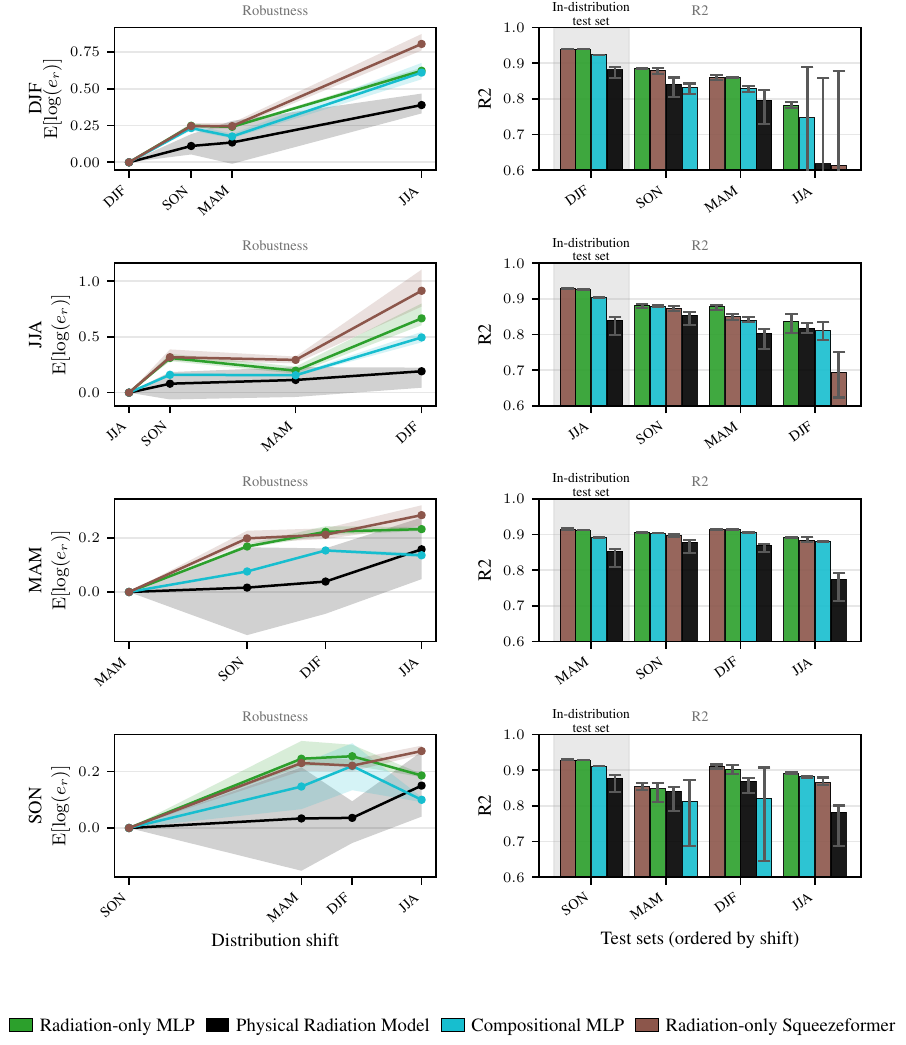}
    \caption{Physically grounded decomposition consistently improves robustness to seasonal shift without sacrificing in-distribution skill, across all training seasons. Each row corresponds to a different seasonal training group (JJA, DJF, SON, MAM) from the NH mid-latitude seasons benchmark; models are evaluated on all four seasonal test sets in every row. Left: Mean log error ratio $\mathbb{E}[\log(e_r)]$ against covariate shift measured by energy distance — a flat line indicates robustness to shift, a rising line indicates degradation; shaded bands show the min–max range across 10 random seeds. Right: $R^2$ on each test set ordered by increasing covariate shift; the shaded column marks the in-distribution test set; error bars show the seed range. Across all training seasons, the compositional MLP improves robustness over all ML baselines while substantially exceeding the physical model's in-distribution $R^2$, confirming that the results reported in Figure~\ref{fig/composing} are not specific to summer training.}
    \label{fig/composing_full}
\end{figure}

Figure~\ref{fig/composing_full} extends the analysis across all four seasonal training groups. We note that hyperparameter search was conducted on JJA only; all other groups use the same architecture and optimisation settings, making this a conservative test of generalisation across training conditions. Nevertheless, the pattern observed in Figure~\ref{fig/composing} holds consistently: the physical model is the most robust to seasonal shift in every row but achieves the lowest in-distribution $R^2$, while ML baselines are more skilful in-distribution but degrade more steeply. The compositional MLP sits between these extremes across all training groups — improving robustness over the ML baselines and recovering substantial in-distribution skill relative to the physical model — suggesting that the benefit of structured decomposition is not specific to any single season.

\subsection{Implications}

This experiment demonstrates that physically motivated decompositions can serve as effective elementary components for compositional generalisation in climate emulation. While we focus here on radiation, the principle that structured decomposition aids OOD robustness is likely to extend to other climate processes, and we leave this exploration to future work.

A central remaining challenge is that our compositional MLP relies on domain knowledge to identify both the relevant partitions and the input variables for each expert. Automating this — learning how to decompose the climate system and which variables matter for each component — is a substantially harder problem. It connects directly to active research in representation learning and causal discovery \cite{lei_spartan_2024, scholkopf2021toward, baumgartner2026disentangling, yao2024marrying}, both of which are beginning to be applied to weather and climate data \cite{iglesias-suarez_causally-informed_2024, beucler_towards_2020, macmillan2025towards, liu2025cera, brouillard2026learning}. Progress on these fronts offers a clear path towards ML emulators that are simultaneously skilful, stable, and robust across the range of conditions relevant to climate projection.

This result also surfaces an important open question for climate emulation: when deploying ML models in unknown future scenarios, is it preferable to prioritise higher in-distribution skill that degrades under shift, or lower skill that remains stable? This is further complicated by cases such as the radiation-only MLP, which, despite steeper degradation, remains more skilful than more robust models across all test sets. We view resolving this trade-off as an important direction for the field.
\newpage

\section{Limitations}
\label{section/appendix/limitations}

Our analysis has several limitations. First, in studying climate shift, we consider only reduced variables at monthly timesteps that are spatially averaged. A more fine-grained evaluation, for example at higher temporal resolution and with greater spatial detail, could reveal location- and time-specific sensitivities to climate change that are not captured here.

Second, when using seasonal shift as a proxy for climate change, we restrict our analysis to emulating a single process at a specific timestep. While emergent constraints provide additional theoretical support for this choice, the extent to which these findings generalise across processes and temporal scales remains unclear.

Third, we find that the energy distance metric can mask shift signals, which may limit the sensitivity of our evaluation. Alternative or complementary metrics could provide a more complete picture of distributional changes.

Finally, we apply our compositional modelling approach only to the emulation of radiation processes. As such, its broader applicability to other components or processes within climate models is not evaluated.

\section{Compute}
\label{appendix/section/compute}

ERA5 experiments were run on a single NVIDIA L40 GPU, with a single MLP variant taking less than five minutes to train. ClimSim experiments were conducted on an NVIDIA A100 GPU, with the longest training time taking approximately 1 hour and 30 minutes, using 64 GB of memory and 2 CPUs. The models and the energy distance metric both have low computational cost and are efficient to run on GPUs.

% \clearpage
% \input{checklist.tex}

\end{document}